\documentclass[10pt,twocolumn,letterpaper]{article}
\pdfoutput=1
\usepackage{iccv}
\usepackage{times}
\usepackage{epsfig}
\usepackage{graphicx}
\usepackage{amsmath}
\usepackage{amssymb}
\usepackage{booktabs}

\usepackage[breaklinks=true,bookmarks=false]{hyperref}

\iccvfinalcopy 

\def\iccvPaperID{0876} 

\ificcvfinal\fi
\begin{document}

\title{3D-LaneNet: End-to-End 3D Multiple Lane Detection}

\author{Noa Garnett \hspace{1cm} Rafi Cohen \hspace{1cm} Tomer Pe'er \hspace{1cm} Roee Lahav\hspace{1cm}  Dan Levi\\
	General Motors Israel\\
	HaMada St. 7, Herzlya, Israel\\
	{\tt\small \{noa.garnett,rafi.cohen,tomer.peer,roee.lahav,dan.levi\}@gm.com}
}

\maketitle
\ificcvfinal\thispagestyle{empty}\fi

\begin{abstract}
	We introduce a network that directly predicts the 3D layout of lanes in a road scene from a single image. This work marks a first attempt to address this task with on-board sensing without assuming a known constant lane width or relying on pre-mapped environments. Our network architecture, 3D-LaneNet, applies two new concepts: intra-network inverse-perspective mapping (IPM) and anchor-based lane representation. The intra-network IPM projection facilitates a dual-representation information flow in both regular image-view and top-view. An anchor-per-column output representation enables our end-to-end approach which replaces common heuristics such as clustering and outlier rejection, casting lane estimation as an object detection problem. In addition, our approach explicitly handles complex situations such as lane merges and splits. Results are shown on two new 3D lane datasets, a synthetic and a real one. For comparison with existing methods, we test our approach on the image-only tuSimple lane detection benchmark, achieving performance competitive with state-of-the-art.
\end{abstract}

\vspace{-0.7cm}

\section{Introduction}


3D lane detection, comprising of an accurate estimation of the 3D position of the drivable lanes relative to the host vehicle, is a crucial enabler for autonomous driving. Two complementary technological solutions exist: loading pre-mapped lanes generated off-line~\cite{Urmson_boss} and perception-based real-time lane detection~\cite{aharon}. The off-line solution is geometrically accurate given precise host localization (in map coordinates) but complex to deploy and maintain. The most common perception-based solution uses a monocular camera as the primary sensor for solving the task. Existing camera-based methods detect lanes in the image domain and then project them to the 3D world by assuming a flat ground~\cite{aharon}, leading to inaccuracy not only in the elevation but also in the lane \textit{curvature} when the assumption is violated. Inspired by recent success of convolutional neural networks (CNNs) in monocular depth estimation~\cite{Liu_Depth}, we propose instead to directly detect lanes in 3D. More formally, given a single image taken from a front-facing camera, the task is to output a set of 3D curves in camera coordinates, each describing either a lane delimiter or a lane centerline.

\begin{figure}[ht]
	\begin{center}
		\includegraphics[width=\linewidth]{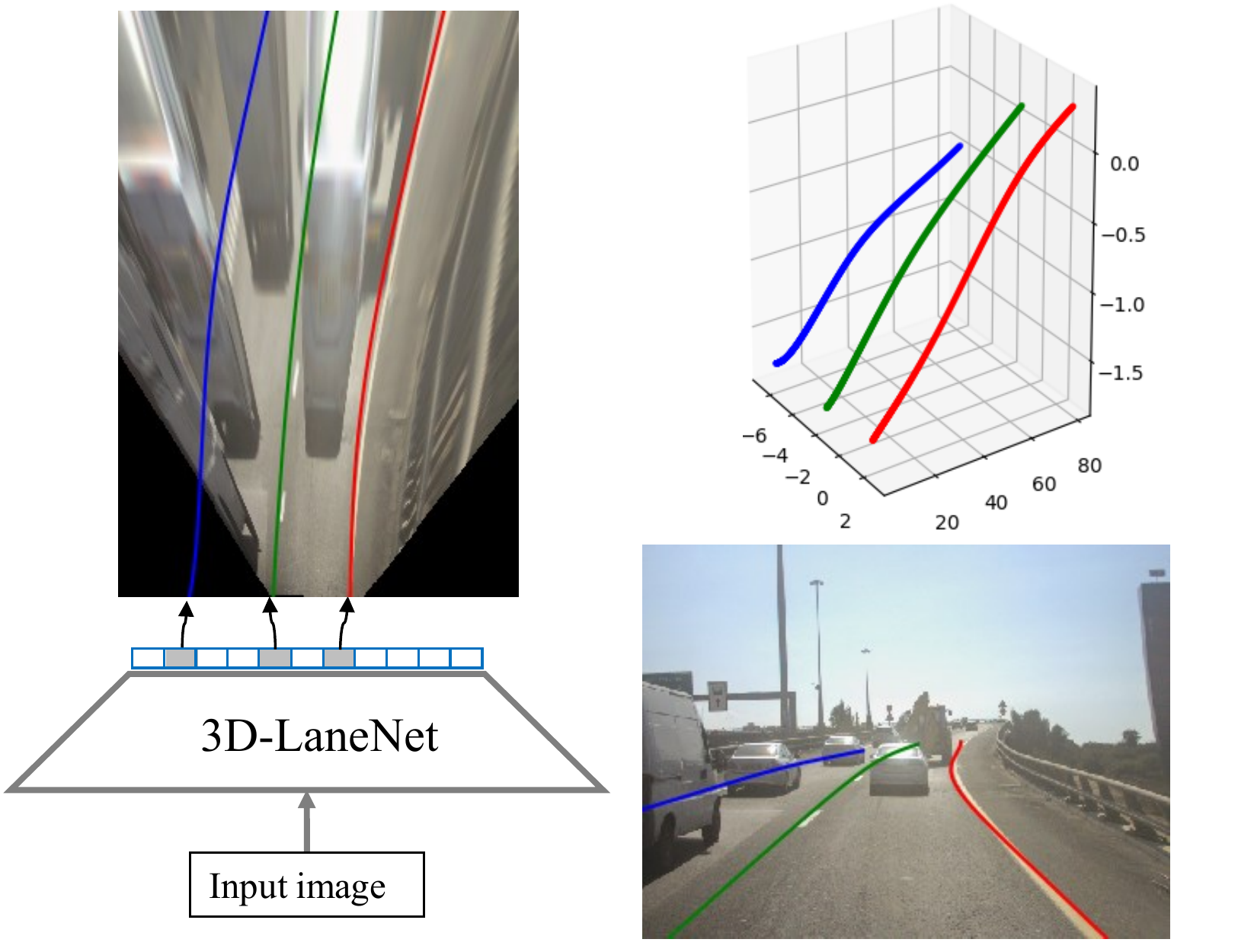}
	\end{center}	
	\caption{\textbf{End-to-end approach illustrated}. \textbf{Left:} Output represented in top view. \textbf{Top-right: }result visualized in 3D. \textbf{Bottom-right:} result projected to original input image}
	\label{fig:tusimple}
\end{figure}

3D-LaneNet, our proposed solution, is a deep CNN that performs 3D lane detection. The network, trained end-to-end, outputs in each longitudinal road slice, the confidence that a lane passes through the slice and its 3D curve in camera coordinates. Our approach is schematically illustrated in Figure~\ref{fig:tusimple}. Our direct, single-shot approach avoids post-processing used in existing methods such as clustering and outlier rejection. The network's backbone is based on a novel dual pathway architecture that uses several in-network projections of feature maps to virtual bird-eye-view. This dual representation gives the network an enhanced ability to infer 3D in a road scene, and may possibly be used for other tasks requiring this ability (e.g. 3D car detection). The output is represented by a new column-based anchor encoding which makes the network horizontally invariant and enables the end-to-end approach. Each output is associated to an anchor in analogy to single-shot, anchor-based object detection methods such as SSD~\cite{SSD} and YOLO~\cite{YOLO}. Effectively, our approach casts the problem as an object detection one, in which each lane entity is an object, and its 3D curve model is estimated just like bounding box for an object.

\begin{figure*}[th]
	\begin{center}
		\begin{tabular}{c|c|c}
			\includegraphics[trim={2.2cm 7.4cm 2.15cm 2.8cm},clip=true,width=0.3\linewidth]{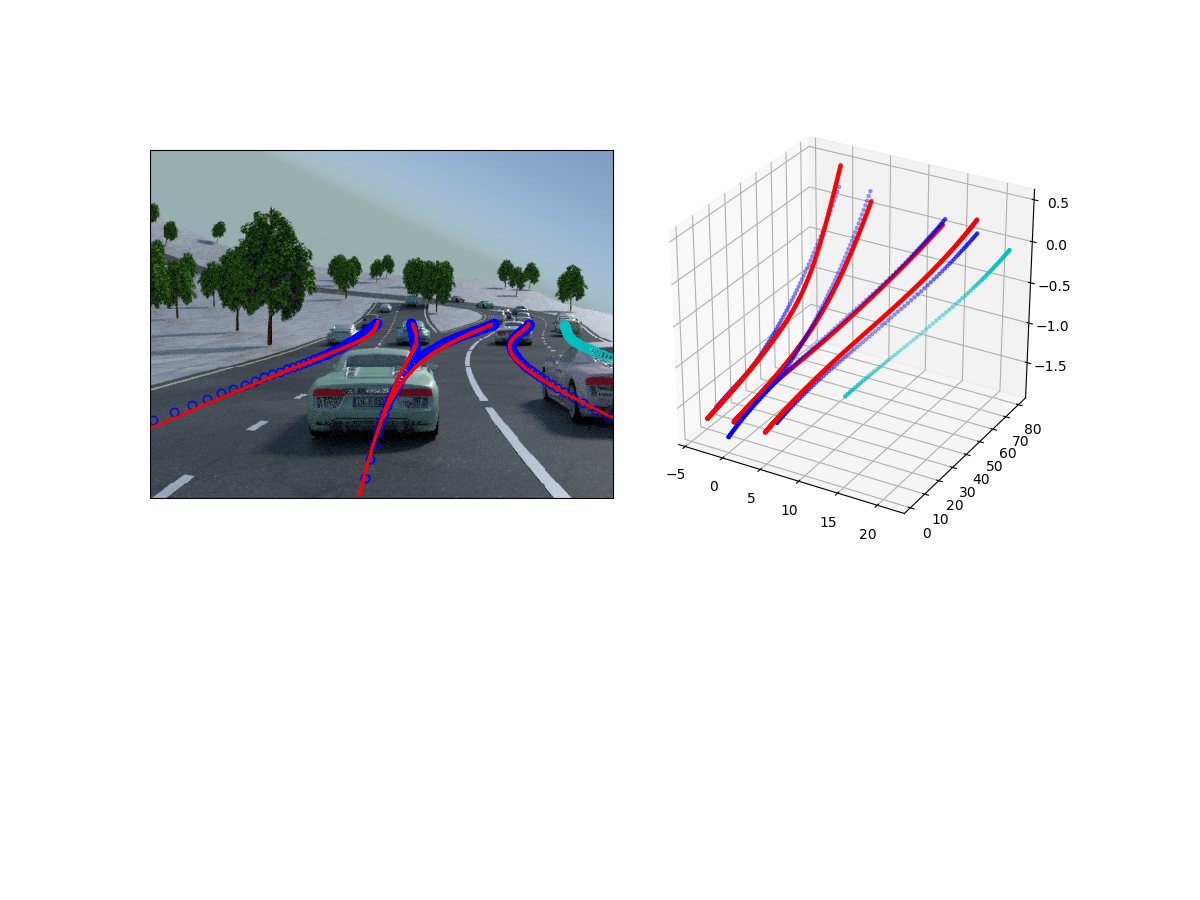} &
			\includegraphics[trim={2.2cm 7.4cm 2.15cm 2.8cm},clip=true,width=0.3\linewidth]{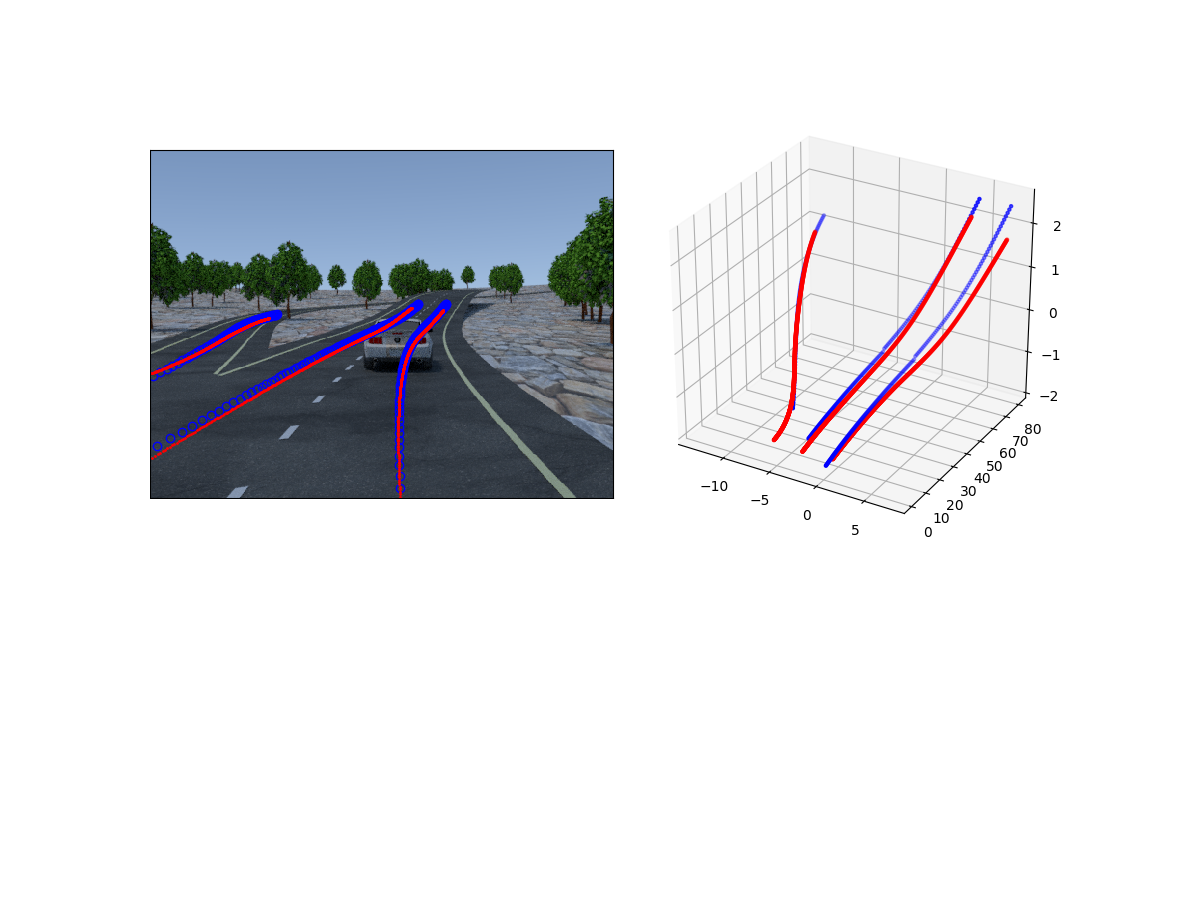} &
			\includegraphics[trim={2.2cm 9.4cm 5.15cm 2.8cm},clip=true,width=0.3\linewidth]{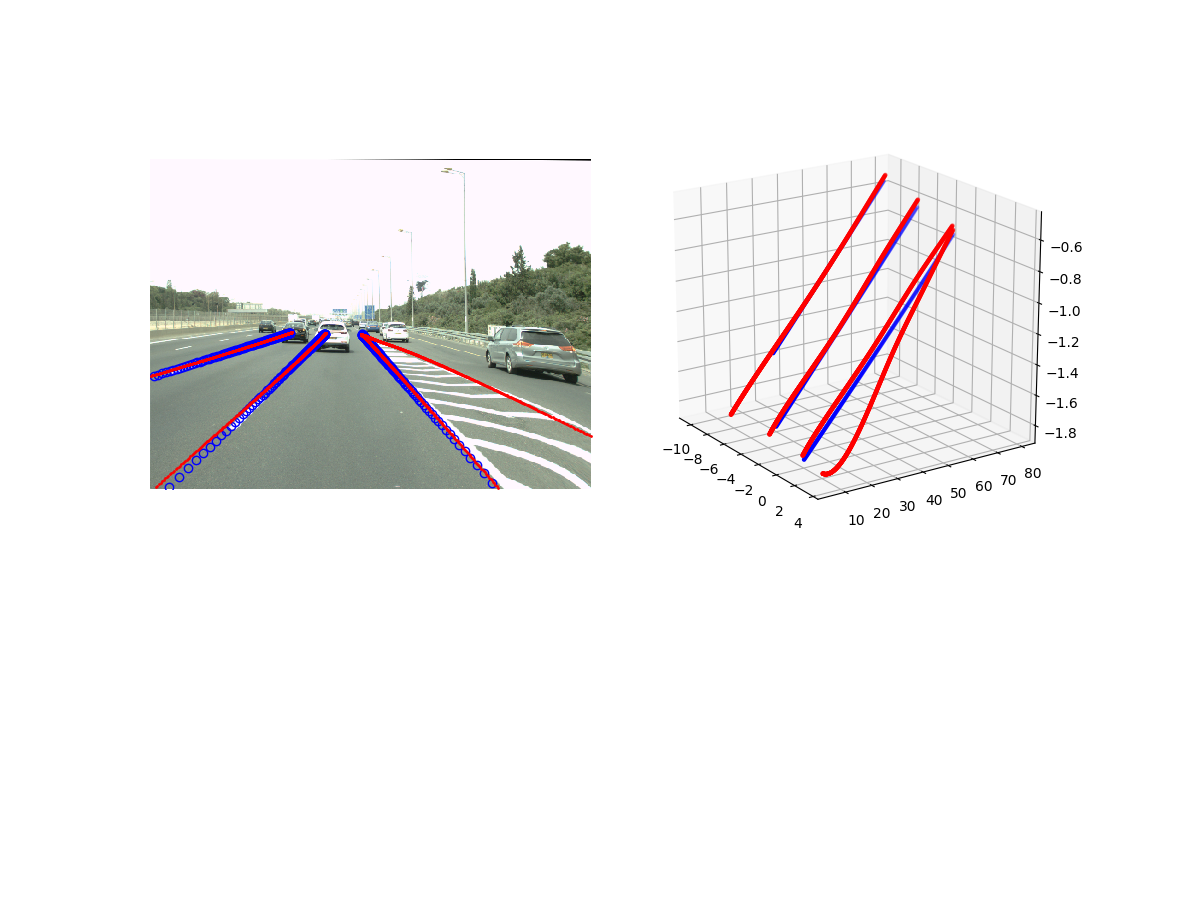} \\
			
			\includegraphics[trim={2.2cm 7.4cm 2.15cm 2.8cm},clip=true,width=0.3\linewidth]{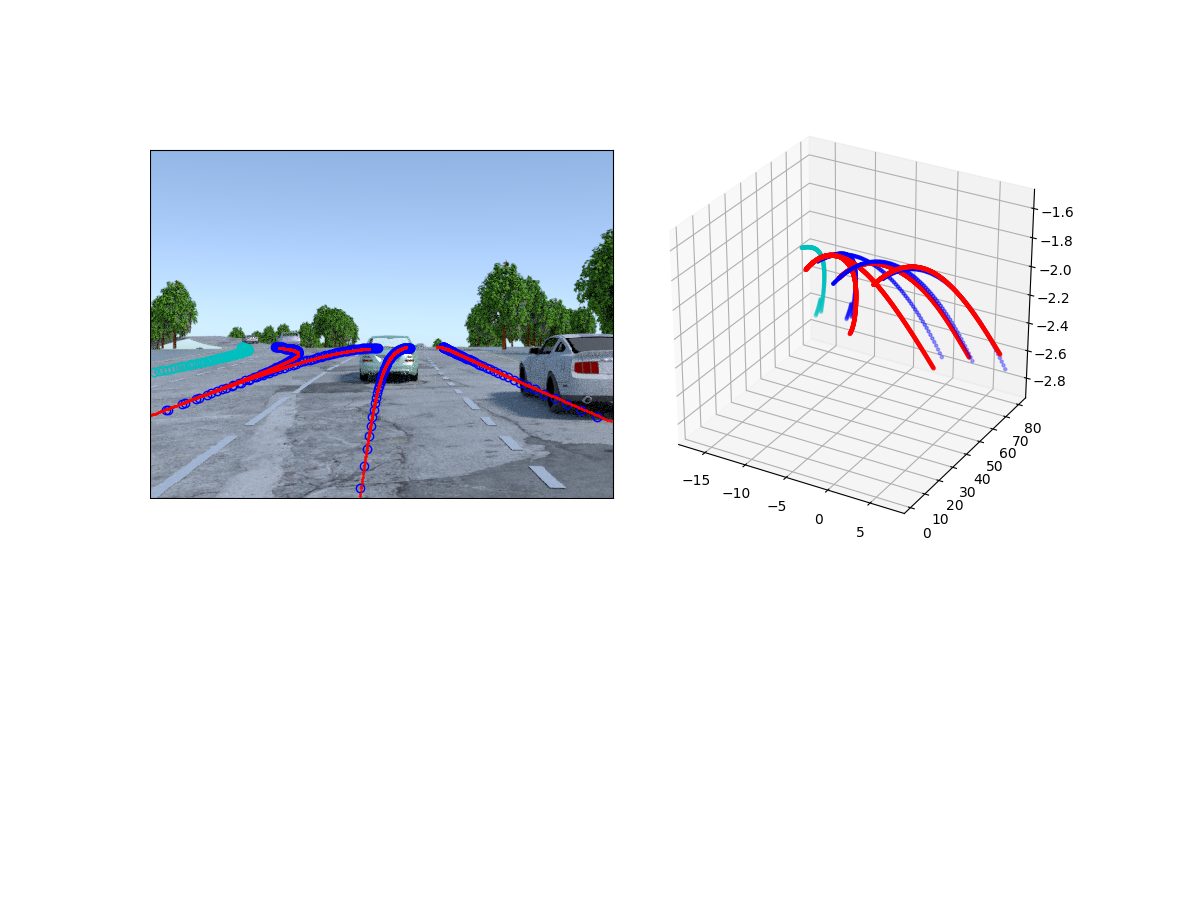} &
			\includegraphics[trim={2.2cm 7.4cm 2.15cm 2.8cm},clip=true,width=0.3\linewidth]{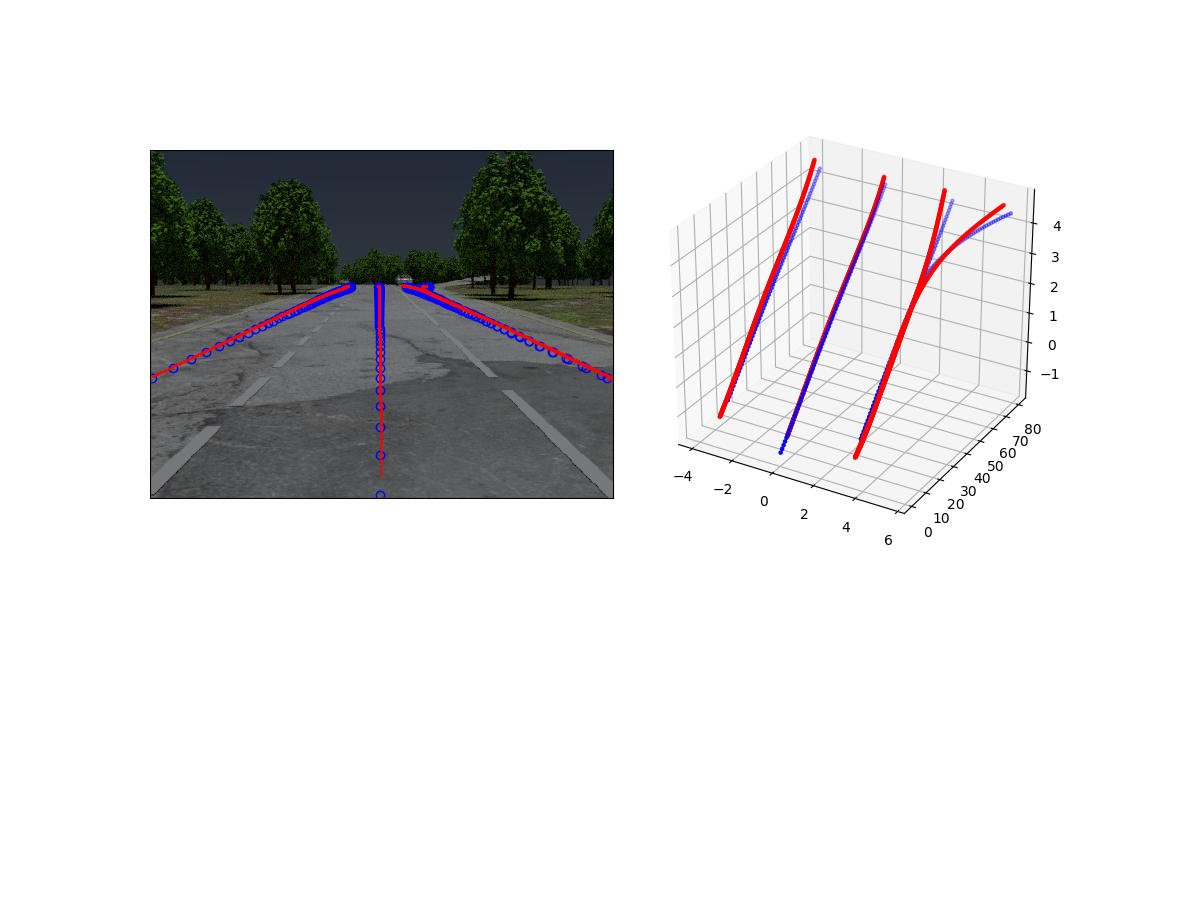} &
			\includegraphics[trim={2.2cm 9.4cm 5.15cm 2.8cm},clip=true,width=0.3\linewidth]{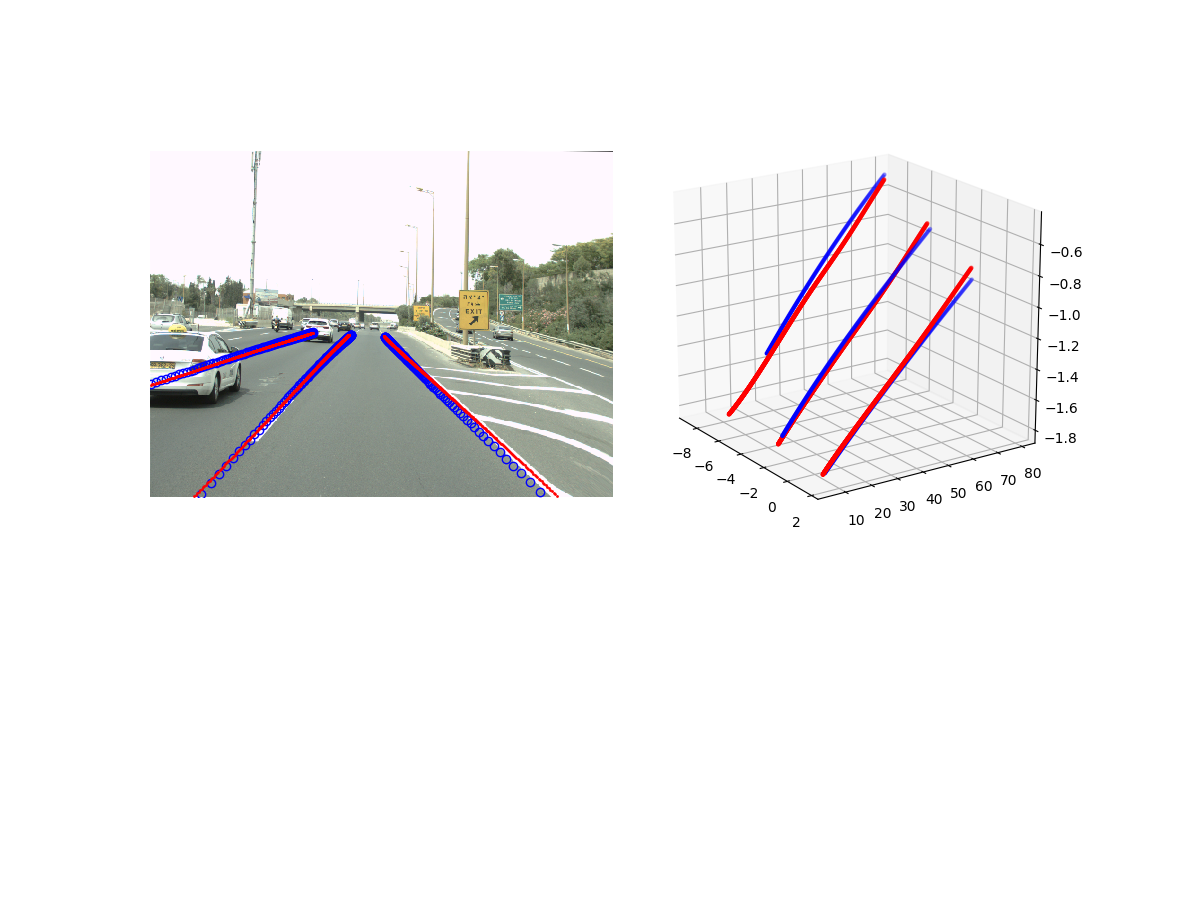} \\
			
			\includegraphics[trim={2.2cm 7.4cm 2.15cm 2.8cm},clip=true,width=0.3\linewidth]{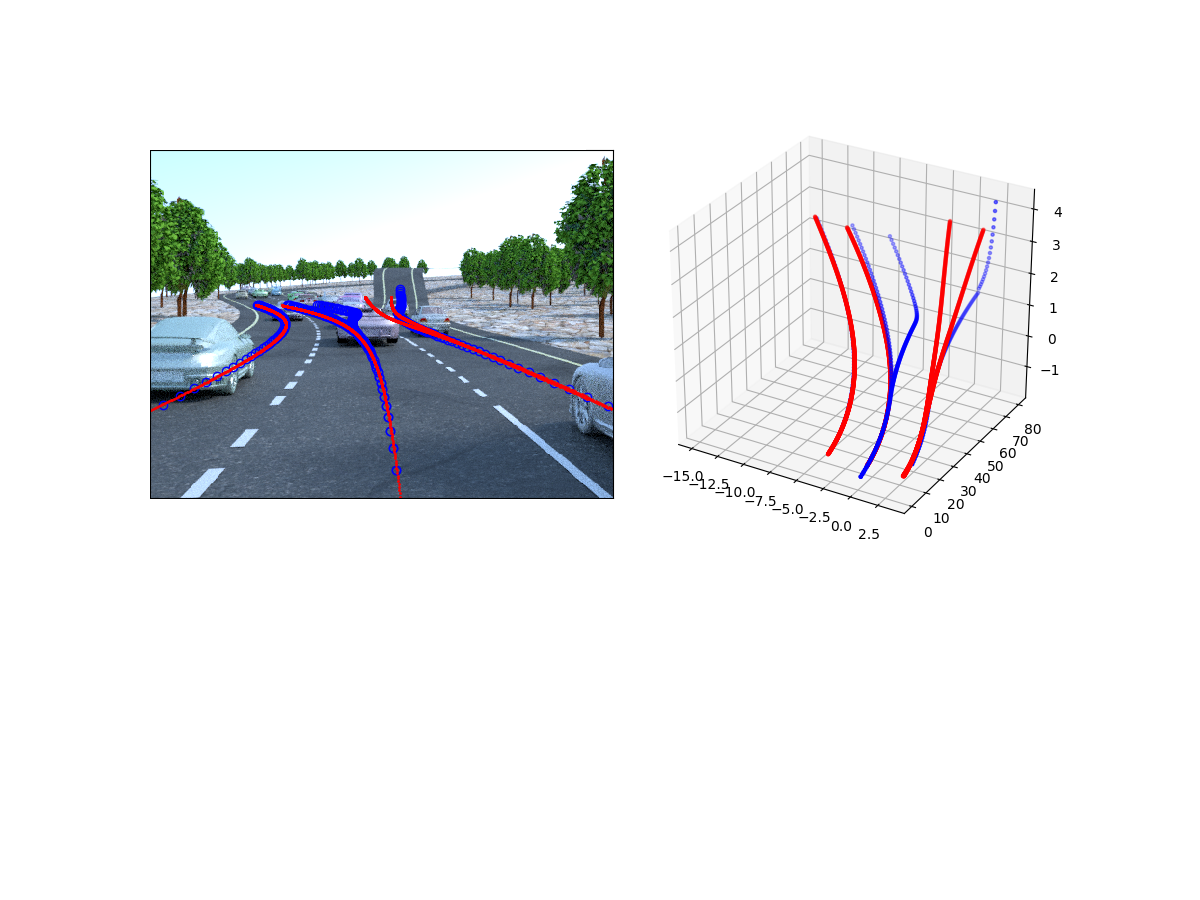} &
			\includegraphics[trim={2.2cm 7.4cm 2.15cm 	2.8cm},clip=true,width=0.3\linewidth]{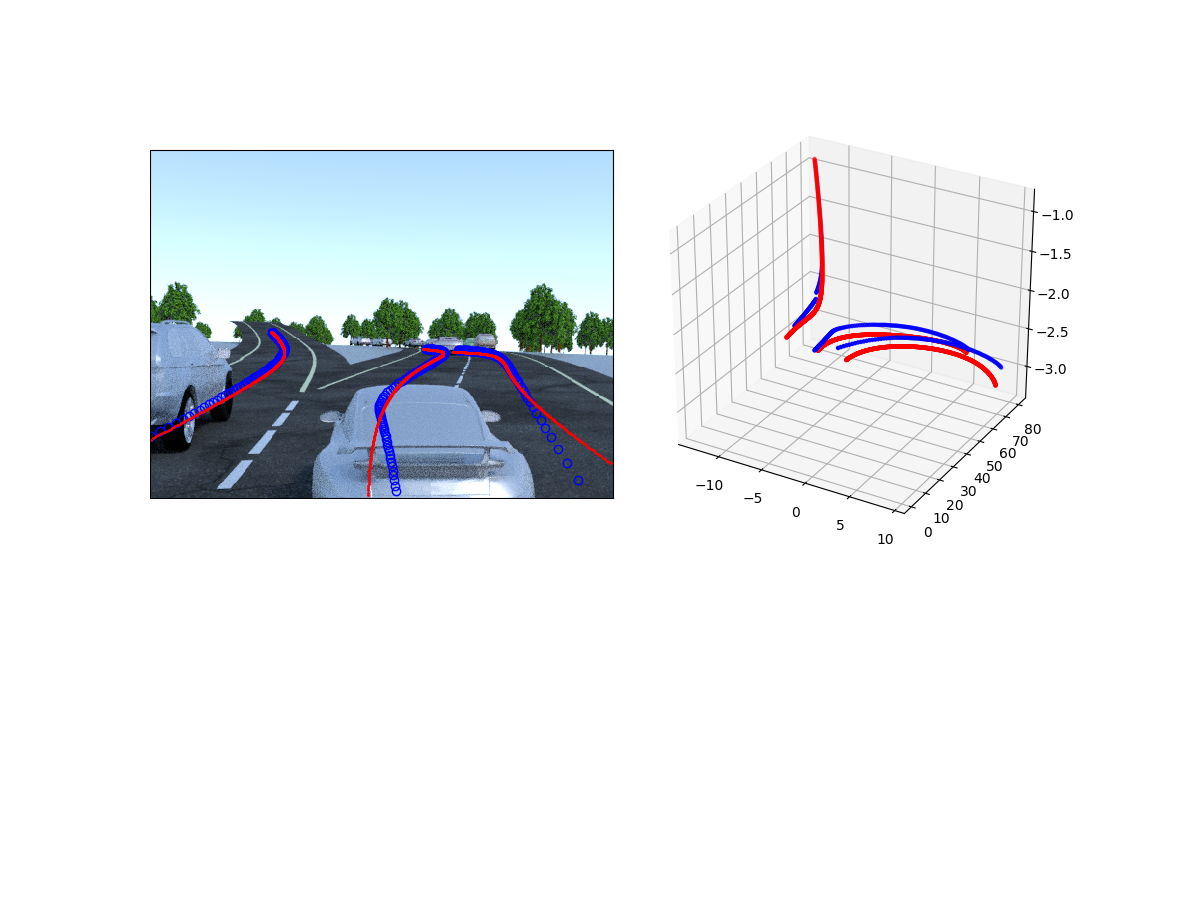} &
			\includegraphics[trim={2.2cm 9.4cm 5.15cm 2.8cm},clip=true,width=0.3\linewidth]{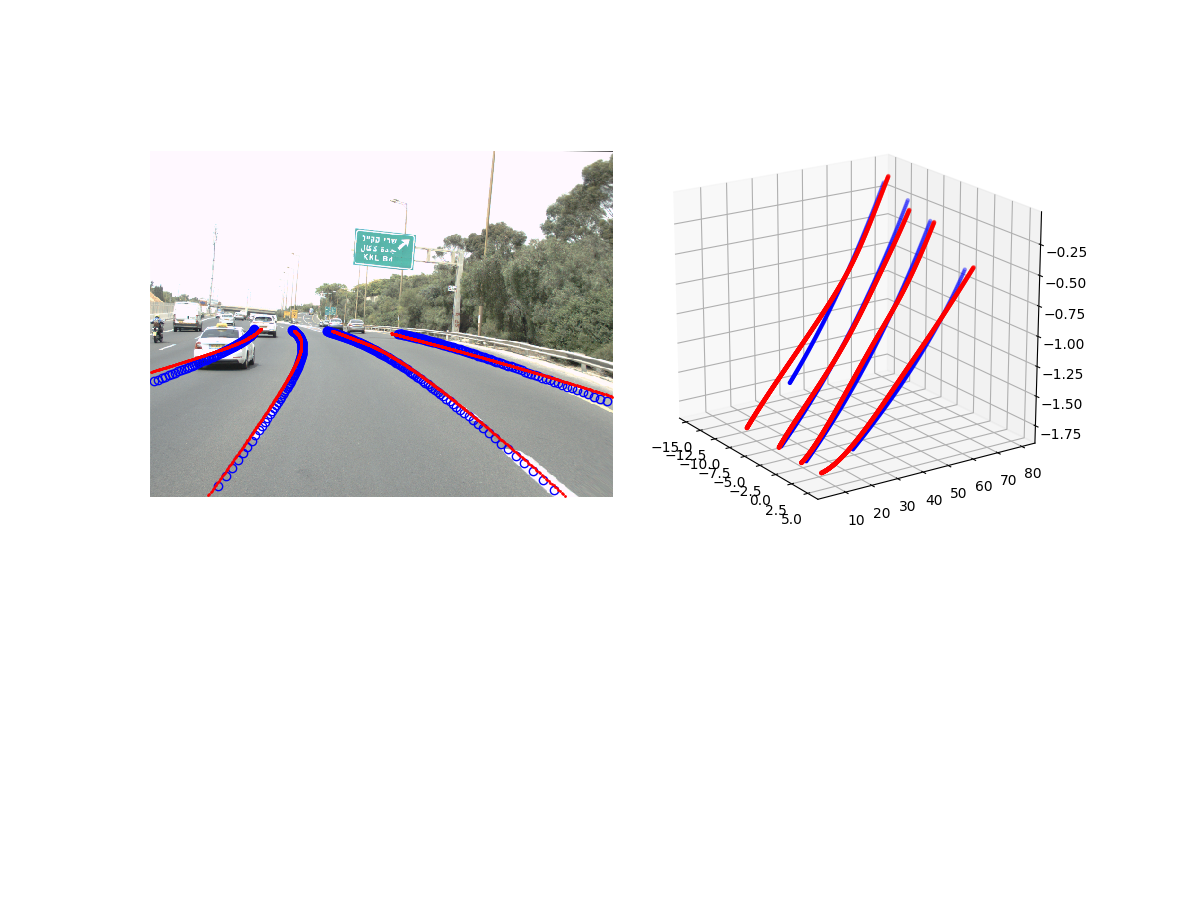} \\
			
		\end{tabular} 
	\end{center}
	\caption{\textbf{Result visualization on test images. Centerline detection on } \textit{synthetic-3D-lanes} \textbf{examples (Left and middle columns) and Delimiter detection on } \textit{3D-lanes} \textbf{real image examples (Right column)}.  Detections with confidence $>0.5$ are shown. Ground truth \emph{\color{blue}(blue)} and method result  \emph{\color{red}(red)} shown in each image alongside a 3D visualization. Note that 3D axes are scene adaptive. Ignored lanes are marked in \emph{\color{cyan}cyan}. The leftmost bottom example shows a failure in correctly assigning a lane split, probably caused by occlusion.}
	\label{fig:results}
	
\end{figure*}


We validate our approach on three different datasets. The primary dataset used to develop the approach is a new computer-graphics dataset, \textit{synthetic-3D-lanes}\footnote{https://sites.google.com/view/danlevi/3dlanes} providing full access to the exact 3D position of each lane element. While several driving simulators exist~\cite{CARLA, AirSim}, they are not focused on the 3D lane detection task, and are limited in the variability of the relevant scene properties (e.g. lane curvature). Our main accomplishment in this domain, is the ability to randomly generate road segments with highly variable 3D shapes and lane topology. We therefore used it as the primary dataset for evaluation and ablation study. To validate our approach on real-world images we collected an additional dataset, \textit{3D-lanes}, from a vehicle-mounted front camera. 3D Lane annotation was accomplished using a Lidar scanner in a semi-manual annotation scheme. Finally, to compare with state-of-the-art lane detection methods, which only operate in the image domain, we adapted our method to this task, and demonstrated end-to-end image-only lane detection. This image-only version is tested on the tuSimple dataset~\cite{tu_simple}, reaching results competitive to state-of-the-art without the common post-processing techniques. To summarize, our main contributions are:


\begin{itemize}\addtolength{\itemsep}{-0.5\baselineskip}
	\item Introduction of a novel problem: single-frame 3D lane detection without geometric assumptions, along with new evaluation metrics
	\item A novel dual-pathway architecture deploying intra-network feature map IPM projections 
	\item An new anchor-based output representation of lanes enabling a direct, end-to-end trained network, for both 3D and image-based lane detection.
	\item A methodology for generating randomly synthetic examples with variation in lane topology (i.e. number of lanes, merges, splits) and 3D shape.
\end{itemize}

\section{Related Work}

Traditional lane detection systems (e.g.~\cite{Gopalan}) combine low-level operations such as directional filters with high-level heuristics such as the Hough transform to detect continuous lanes in the image. A common pipeline includes 4 stages: local lane-feature extraction (1), lane model fitting (2), image-to-world correspondence (3) and temporal aggregation (4). Bar-Hillel \textit{et al.}~\cite{aharon} provide this modular decomposition alongside a detailed overview of traditional systems. In recent years, the \textit{local feature extraction stage} is performed by applying one or more CNNs to the image, but the overall pipeline remains very similar and the latter post processing stages remain. 

Initially, CNNs were used to improve feature extraction by either enhancing the edge map (Kim and Lee~\cite{kim_lee}) or classifying candidate patches (He \textit{et al.}~\cite{He}). Huval \textit{et al.}~\cite{huval} detects local lane line segments with an object detection CNN. VPGNet (Lee \textit{et al.}~\cite{lee}), follows a similar concept and additionally detects other road markings and the vanishing point to improve lane detection. Kim and Park~\cite{kim_park} re-formulate the local-feature extraction stage as a semantic-segmentation problem, with two classes corresponding to the left and right lane delimiters, extending the reach of the network to perform clustering. However, a world-coordinate lane model must still be fitted to each cluster, and multiple lanes are not handled. Neven \textit{et al.}~\cite{neven} make an attempt towards end-to-end multi-lane detection, by training a CNN not only to create a binary lane pixel mask but also a feature embedding used for clustering lane points. Ghafoorian \textit{et al.}~\cite{Ghafoorian} propose applying a generative adversarial network to make the semantic segmentation network output more realistic in the context of lane detection. Several works (e.g. Meyer \textit{et al.}~\cite{Meyer}, Oliveira \textit{et al.}~\cite{Oliveira}) are built on a similar approach in which the host and possibly adjacent lanes are the semantic classes (lane interior rather than the lane delimiters).


As opposed to all presented methods, 3D-LaneNet unifies the first three stages of the common pipeline by providing a full multi-lane representation in 3D world coordinates directly from the image in a single feed-forward pass. In addition, previous methods use the flat ground assumption for the image-to-world correspondence while our method fully estimates the parametrized 3D curves defining the lanes. Only a few methods directly address 3D lane estimation: ~\cite{Nedevschi2004}, using stereo, and ~\cite{Xiong2018, Coulombeau_Laurgeau2002} which follow a multi-view geometry approach and assume a known constant road / lane width to solve the depth ambiguity. Instead, we use a data driven approach and make no geometric assumptions.

Inverse perspective mapping (IPM) generates a virtual top-view (sometimes called bird-eye-view) of the scene from a camera-view as in the example in Figure~\ref{fig:tusimple}. It was introduced in the context of obstacle detection by Mallot \textit{et al.}~\cite{Mallot} and first used for lane detection by Pomerleau~\cite{Pomerleau}. IPM has since been extensively used for lane detection (e.g.~\cite{Borkar,Aly}) since lanes are ordinarily parallel in this view and their curvature can be accurately fitted with low-order polynomials. In addition, removing the perspective effects causes lane markings to look similar (except for blurring effects) regardless of their distance from the camera. Most recently He \textit{et al.}~\cite{He} introduced a ``Dual-view CNN'' which is composed of two separate sub-networks, each producing a descriptor (one per view) which are then concatenated and applied to candidate image locations. Li \textit{et al.}~\cite{Li} use a CNN to detect lane markings along with geometrical attributes, such as local position and orientation, directly on a top-view image which preserves invariance to these properties. In addition they deploy a second, recurrent network, that traverses the image to detect consistent lanes. Neven \textit{et al.}~\cite{neven} use the horizon, predicted in each image by a sub-network (``H-net''), to project the lanes to top-view for improved curve fitting. In contrast to previous work, we exploit both views in a synergistic single network approach.

More generally, we propose the first method that uses an end-to-end trained CNN to directly detect multiple lanes and estimate the 3D curvature for each such lane. We also show that our method is applicable both to centerlines and delimiters with an ability to handle splits and merges as well, without any further post-processing. 

\section{Method}

\begin{figure}[t]
	\begin{center}
		\includegraphics[width=0.79\linewidth]{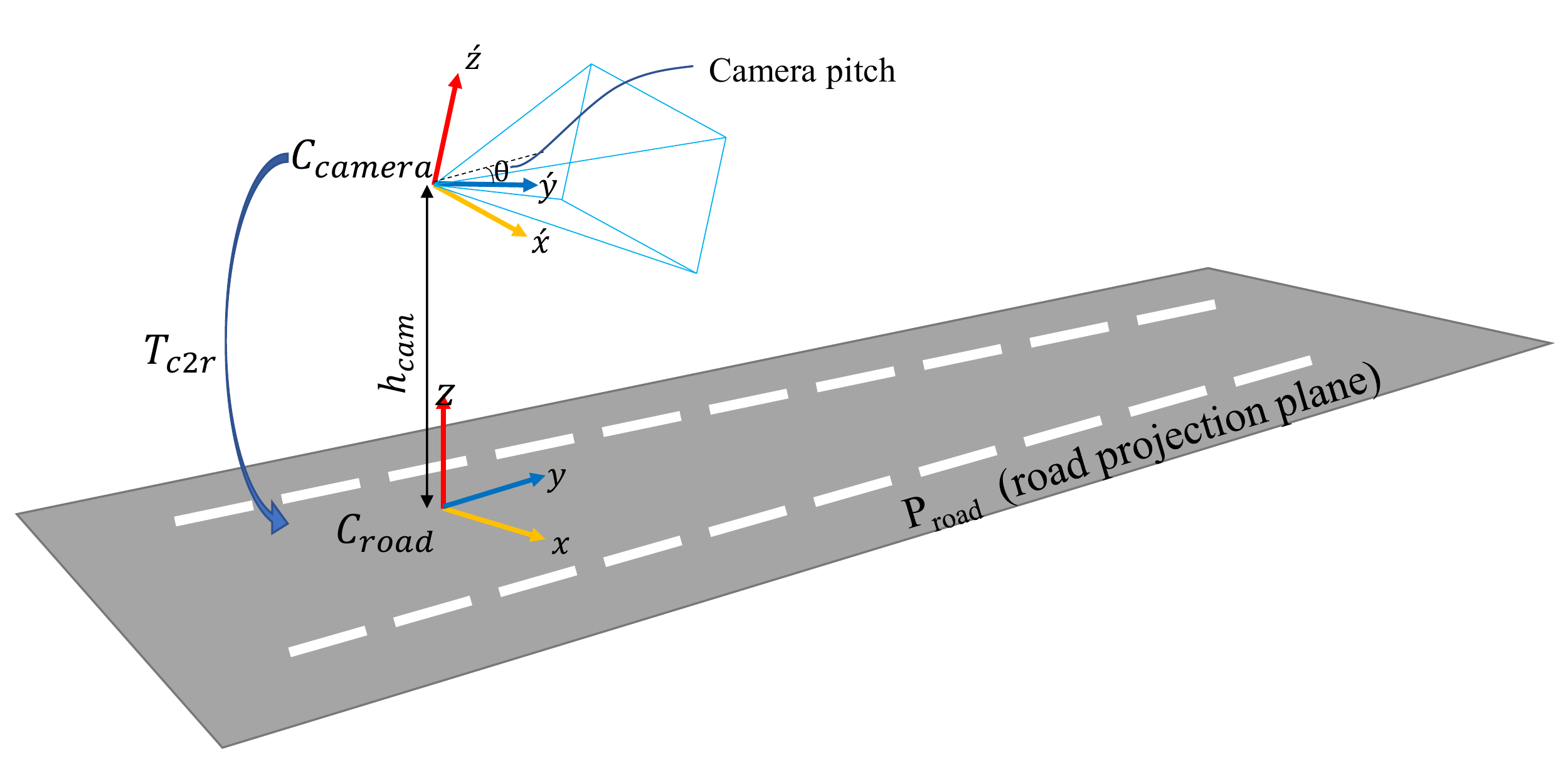}
	\end{center}
	\caption{\textbf{Camera position and road projection plane}}
	\label{fig:geometry}
	
\end{figure}

Our method gets as input a single image taken from a front facing camera mounted on a vehicle as illustrated in Figure~\ref{fig:geometry}. We assume known intrinsic camera parameters $\kappa$ (e.g. focal length, center of projection). We also assume that the camera is installed at zero degrees roll \textit{relative to the local ground plane}. We do not assume a known camera height and pitch since these may change due to vehicle dynamics. The lanes in a road scene can be described both by the set of centerlines $\left\lbrace C_i\right\rbrace_{i=1}^{N_C}$ of each lane and by the set of lane delimiters $\left\lbrace D_i\right\rbrace_{i=1}^{N_D}$ as illustrated in Fig.~\ref{fig:out}. Each such lane entity (centerline or delimiter) is a curve in 3D expressed in camera coordinates ($\mathcal{C}_{camera}$).  The task is to detect either the set of lane centerlines and/or lane delimiters given the image.


\begin{figure*}[t]
	\begin{center}
		\includegraphics[width=\linewidth]{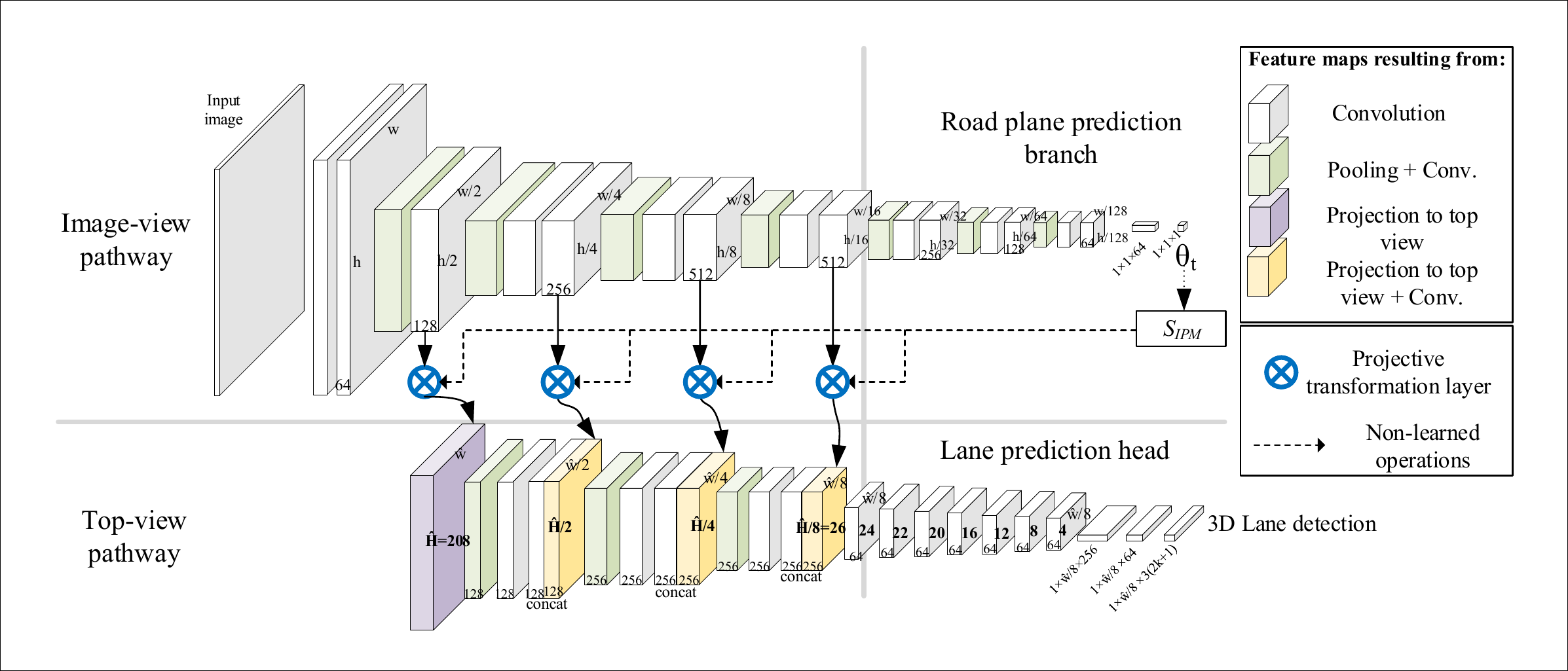} 
	\end{center}
	\caption{\textbf{3D-LaneNet network architecture.}}
	\label{fig:net}
\end{figure*}

\begin{figure}[t]
	\begin{center}
		\includegraphics[width=0.7\linewidth]{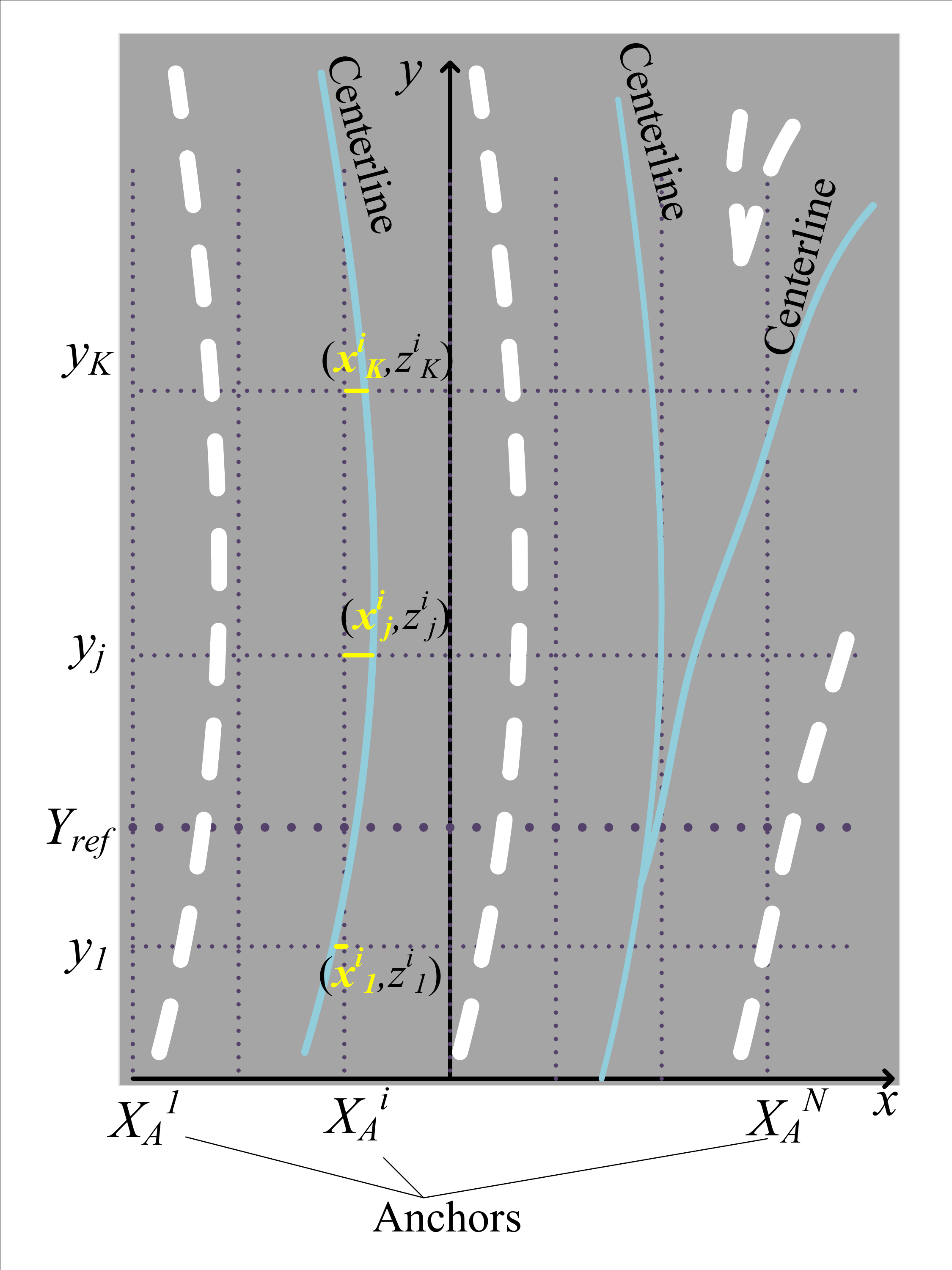}
	\end{center}
	\caption{\textbf{Output representation.} Note that the number of anchors ($N$) equals the output layer width (marked by $\hat{w}/8$ in Fig.~\ref{fig:net}). The geometry representation for one lane \textit{centerline} is shown. Lane \textit{delimiters} (white dashed curves) are similarly represented.}
	\label{fig:out}
\end{figure}

\subsection{Top-view projection}\label{sec:IPM}
We briefly review Inverse Perspective Mapping (IPM). In short, IPM is a homography that warps a front view image to a virtual top view image as depicted in the top-left image of Figure~\ref{fig:tusimple}. It is equivalent to applying a camera rotation homography (view is rotated downwards) followed by an anisotropic scaling~\cite{Hartley2004}. In our implementation we want to ensure that each pixel in the top view image corresponds to a predefined position on the road, \textit{independent} of the camera intrinsics and its pose relative to the road.

See Figure~\ref{fig:geometry} for an illustration of the following definitions. The camera coordinates $\mathcal{C}_{camera}=\left(\acute{x},\acute{y},\acute{z}\right)$  are set such that $\acute{y}$ is the camera viewing direction. Let $P_{road}$ be the plane \textit{tangent} to the local road surface. We define the road coordinates $\mathcal{C}_{road}=\left(x,y,z\right)$ as follows: $z$ direction is the normal to $P_{road}$, $y$ is the projection of $\acute{y}$ onto $P_{road}$ and the origin is the projection of the camera center onto $P_{road}$. Let $T_{c2r}$ be the 6-D.O.F. transformation between $\mathcal{C}_{camera}$ and  $\mathcal{C}_{road}$ (3D translation and 3D rotation). Since we assume zero camera roll, $T_{c2r}$ is uniquely defined by the camera pitch angle $\theta$ and its height above the ground $h_{cam}$. The homography $H_{r2i}:\mathbb{P}^2\mapsto\mathbb{P}^2$, mapping each point on $P_{road}$ to image plane coordinates, is determined by $T_{c2r}$ and $\kappa$ (See ~\cite{Hartley2004}, Section 8.1.1). Finally, the IPM is obtained from $H_{r2i}$ using a fixed set of parameters $IPM_{Params}$  defining the top view region boundaries and an anisotropic scaling from meters to pixels. The top view image is generated using bilinear interpolation defined by a sampling grid $S_{IPM}$. 


\label{sec:dual_block}

\subsection{Network structure}

An overview of the 3D-LaneNet is illustrated in Figure~\ref{fig:net}. Information is processed in two parallel streams or pathways: the image-view pathway and the top-view pathway. We call this the \emph{dual-pathway backbone}. The image-view pathway processes and preserves information from the image while the top-view pathway provides the features with translation invariance and is used to predict the 3D lane detection output. The architecture for the image-view pathway is based on VGG16~\cite{VGG} while the top-view pathway is similarly structured. Information flows to the top-view pathway through four \textit{projective transformation} layers as follows. 


\subsection{The projective transformation layer}

A main building block in our architecture is the \emph{projective transformation layer} marked in blue in Fig.~\ref{fig:net}. This layer is a specific realization, with slight variations, of the \textit{spatial transformer  module}~\cite{spatial_transform}. It performs a differentiable sampling of input feature maps corresponding spatially to the \textit{image plane}, to output feature maps corresponding spatially to a \textit{virtual top view} of the scene while preserving the \# of channels. The differential sampling is achieved through a grid generated as described in Sec. ~\ref{sec:IPM}, using a IPM predicted by the \textit{Road projection prediction branch} as described in the next section. The resulting projected feature maps, except for the first set, are concatenated to downstream feature maps from the top-view pathway. A subsequent neuron, operating on the concatenated feature maps combines the following two desirable properties for lane detection. First, translational invariance in the top-view plane. This is valuable since in the top view lanes have similar appearance and geometry across the space. Second, preservation of a dual information context - in both image and top view. The additional image-view context encodes information which is not present in the top view such as fences, skyline and trees which are crucial for deciphering the 3D structure of the scene. Particularly, in the far range, the image-view context is much richer in visual information, and represents a much larger actual area compared to the top view.

\subsubsection{Road projection prediction branch}
\label{horizon_estimation}
The first intermediate output of the image-view pathway network is an estimation of  the ``road projection plane'' $P_{road}$. Essentially, this branch predicts $T_{c2r}$, the camera ($\mathcal{C}_{camera}$) to road ($\mathcal{C}_{road}$) transformation. It is trained in a supervised manner. $T_{c2r}$ determines the top-view homography $H_{r2i}$ and sampling grid $S_{IPM}$ as explained in Section~\ref{sec:IPM}, and is therefore needed for the feed-forward step of the top-view pathway. At inference time, it is used also to translate the network output which is expressed in $\mathcal{C}_{road}$, back to $\mathcal{C}_{camera}$. As described in Section~\ref{sec:IPM}, $T_{c2r}$ is defined in our case by the camera height $h_{cam}$ and pitch $\theta$, and therefore these are the two outputs of this branch.

\subsubsection{Lane prediction head}

At the heart of our end-to-end approach lies the \textbf{anchor-based lane representation}. Inspired by object detection, we use anchors to define lane candidates and a refined geometric representation to describe the precise 3D lane shape for each anchor. The output coordinate system is the estimation of $\mathcal{C}_{road}$ determined by $h_{cam}$, $\theta$. Our anchors correspond to longitudinal lines in this coordinate system and the refined lane geometry to 3D points relative to the respective anchor. As illustrated in Figure~\ref{fig:out}, we define the anchors by equally spaced vertical (longitudinal) lines in $x$-positions $\left\lbrace X_A^i \right\rbrace _{i=1}^{N}$. Per anchor $X_A^i$, a $3D$ lane is represented by $2\cdot  K$ output neurons activation $\left(\mathbf{x}^i,\mathbf{z}^i\right)=\left\lbrace \left(x_j^i,z_j^i\right)\right\rbrace_{j=1}^K$, which together with a fixed vector of $K$ predefined $y$ positions ($\mathbf{y}=\left\lbrace y_j \right\rbrace_{j=1}^K$) define a set of $3D$ lane points. The values $x_j^i$ are horizontal offsets \emph{relative to the anchor position $X_A^i$}. Meaning, the output $\left(x_j^i,z_j^i\right)$ represents the point $\left(x_j^i+X_A^i,y_j,z_j^i\right)\in\mathbb{R}^3$, in $C_{road}$ coordinates. In addition, for each anchor $i$, we output the confidence $p^i$ that there is a lane associated with the anchor. We use a predefined longitudinal coordinate $Y_{ref}$ for the association. The anchor $X_A^i$ associated to a lane is the one closest to the x-coordinate of the lane at $y=Y_{ref}$.

Per anchor, the network outputs up to \emph{three} types ($t$) of lane descriptors (confidence and geometry), the first two ($c_1,c_2$) represent lane centerlines and the third type ($d$) a lane delimiter. Assigning two possible centerlines per anchor yields the network support for merges and splits which may often result in having the centerlines of two lanes coincide at $Y_{ref}$ and separating at different road positions as in the rightmost example in Figure~\ref{fig:out}. The topology of lane \textit{delimiters} is generally more complicated compared to centerlines and our representation cannot capture all situations (for example the lane delimiters not crossing $y=Y_{ref}$ in Fig.~\ref{fig:out}). The prediction head of the 3D-LaneNet is designed to produce the described output. Through a series of convolutions with no padding in the $y$ dimension, the feature maps are reduced, and finally the prediction layer size is $3\cdot \left(2\cdot K +1\right)\times 1 \times N$ s.t. each column $i\in\left\lbrace 1\ldots N\right\rbrace$ corresponds to a single anchor $X_A^i$. Per anchor, $X_A^i$ and type $t\in \left\lbrace c_1, c_2, d\right\rbrace$ the network output is denoted by $\left(\mathbf{x}_t^i,\mathbf{z}_t^i,p_t^i\right)$. 
The final prediction performs a 1D non-maximal suppression as common in object detection: only lanes which are locally maximal in confidence (compared to the left and right neighbor anchors) are kept. Each remaining lane, represented by a small number ($K$) of 3D points, is translated to a smooth curve using spline interpolation.

\subsection{Training and ground truth association}\label{sec:training}

Given an image example and its corresponding 3D lane curves, $\left\lbrace C_i\right\rbrace_{i=1}^{N_C}$ (centerlines) and $\left\lbrace D_i\right\rbrace_{i=1}^{N_D}$ (delimiters), the training proceeds as follows. First, the ground truth (GT) coordinate system $\mathcal{C}_{road}$ is defined for the local road tangent plane as described in Sec.~\ref{sec:IPM} using the known pitch ($\hat{\theta}$) and camera height ($\hat{h}_{cam}$). Next, each lane curve, projected to the $x-y$ plane of $\mathcal{C}_{road}$, is associated with the closest anchor at $Y_{ref}$. The leftmost lane delimiter and leftmost centerline associated with an anchor are assigned to the $c_1$ and $d$ output types for that anchor. If an additional centerline is associated to the same anchor it is assigned to output type $c_2$. This assignment defines the GT per example in the same format as the output: per anchor $X_A^i$ and type $t$ the associated GT is denoted by $\left(\mathbf{\hat{x}}_t^i,\mathbf{\hat{z}}_t^i,\hat{p}_t^i\right)$, where $\hat{p}_t^i$ is an anchor/type assignment indicator, and the coords in $\mathcal{C}_{road}$.  

Both in training time and in the evaluation, entire lanes are ignored if they do not cross $Y_{ref}$ inside valid top-view image boundaries, and lane points are ignored if occluded by the terrain (i.e. beyond a hill top). The overall loss function of the network is given in Eq.~\ref{eq:loss}. It combines three equally weighed loss terms: lane detection (Cross-entropy-loss), lane geometry and road plane estimation ($L_1$-loss). 

\begin{align}
	\label{eq:loss}
	\begin{split}
		\mathcal{L}=&-\sum_{t\in\left\lbrace c_1,c_2, d\right\rbrace}\sum_{i=1}^N \left(\hat{p}_t^i\log p_t^i + \left(1-\hat{p}_t^i\right)\log\left(1-p_t^i\right)\right) \\
		& + \sum_{t\in\left\lbrace c_1,c_2, d\right\rbrace}\sum_{i=1}^N  \hat{p}_t^i\cdot\left(\left\|\mathbf{x}_t^i-\mathbf{\hat{x}}_t^i \right\|_1   +\left\|\mathbf{z}_t^i-\mathbf{\hat{z}}_t^i \right\|_1\right)  \\ 
		& + \left| \theta -\hat{\theta}\right| + \left| h_{cam} -\hat{h}_{cam}\right|
	\end{split}
\end{align}

\section{Experiments}
Our experimental work is presented as follows. We first present the methodology used for generating a new synthetic dataset \emph{synthetic-3D-lanes}, which is used to derive most of this study conclusions. Next, we introduce the \emph{3D-lanes} dataset generated for validation on real-world imagery. Using a newly proposed evaluation method for 3D lane detection, we then present results on both datasets, including an ablation study carefully examining the contribution of each concept in our overall approach. Finally, we compare an image-only version of 3D-LaneNet to existing state-of-the-art methods on the tuSimple benchmark~\cite{tu_simple}.

		

\begin{figure}[h]
	\begin{center}
		\begin{tabular}{c}
			\begin{tabular}{cc}
			(a)\includegraphics[width=0.47\linewidth]{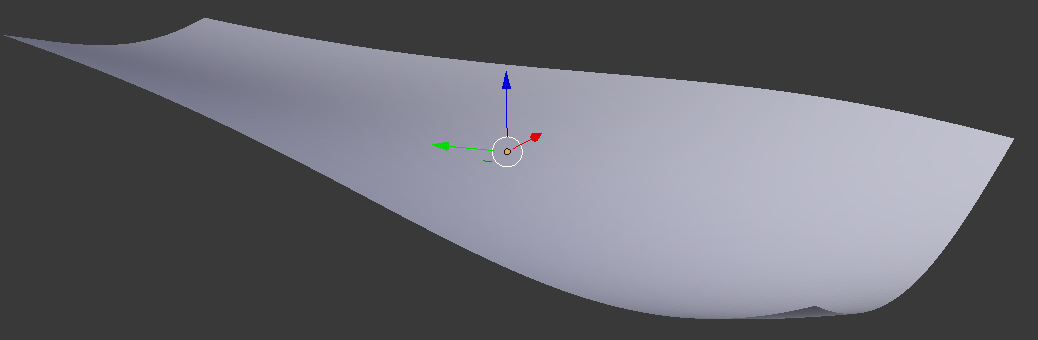} & (b)\includegraphics[width=0.35\linewidth]{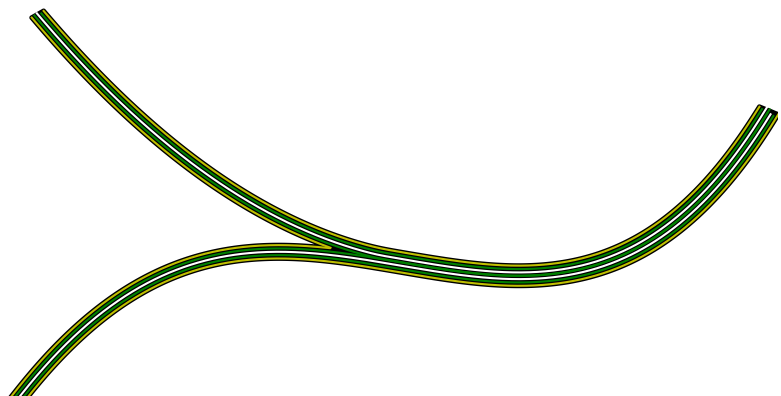}
			\end{tabular}\\
			(c)\includegraphics[width=0.7\linewidth]{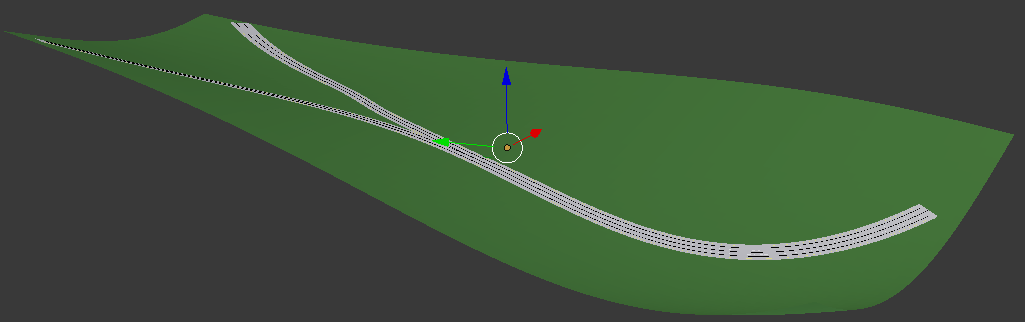}
		\end{tabular}
	\end{center}
	\caption{\textbf{Synthetic scene generation example.} (a) Surface  (b) Road topology and curvature (c) Road on surface}
	\label{fig:synthetic}
\end{figure}

\subsection{Synthetic 3D lane dataset}\label{sec:synthetic}
We generated the \emph{synthetic-3D-lanes} dataset using the open source graphics engine \textit{blender}~\cite{blender}. Our programmatic approach allows us to randomize each of the modeled elements, from the 3D geometry of the scene to object types as illustrated in Figure~\ref{fig:synthetic}. The process of generating \textit{each scene} is composed of the following steps:

\textbf{Terrain 3D.} The terrain is modeled by a Mixture of Gaussians distribution with the number of Gaussians and their parameters randomized. Figure~\ref{fig:synthetic}(a) shows an example of such a terrain.

\textbf{Lane topology.} Number of lanes on the main road is selected. Then we choose if there is a secondary road and the number of lanes in it. Depending on the later direction of the camera in the scene the junction of the secondary road is viewed as a merge or a split.

\textbf{Lane top view geometry.} The geometry of the main road in top view is modeled by a 4\textsuperscript{th} degree polynomial producing mild to extreme curvatures. The junction point for merges / splits as well as the lane width are chosen. This results in a top view lane-level map as shown in Fig.~\ref{fig:synthetic}(b).

\textbf{Lane 3D.} The top-view lane map is placed on the terrain, and secondary roads are lifted to simulate common road topography. Fig.~\ref{fig:synthetic}(c) shows the result of this stage.

\textbf{Terrain and road appearance.} The texture of the road and the terrain are chosen from a set of textures. The type and color of the lane markings is also randomized.

\textbf{Objects.} Cars and trees selected from a set of models are placed in the scene, on and off the road respectively.

\textbf{Scene rendering.} The host vehicle camera is positioned on the main road by choosing its lane and a lateral offset around lane center. The camera height is set randomly between 140cm and 190cm and a downward pitch between 0 and 5 degrees is selected. Finally, illumination is set and the scene is rendered from the camera view. The 3D points of each lane centerline and delimiter are translated to camera coordinates to generate the ground truth.  

Each generated example consists of an image ($360\times480$ pixels) and its associated ground truth: 3D lanes, camera height and pitch. Figure~\ref{fig:results}(Left and middle columns) presents several examples showing the resulting diversity and complexity. The exact parameters used in the random generation process are listed in Appendix I. The generated dataset contains 300K train and 5K test examples. An additional 1K validation set was used for learning rate scheduling and choosing the best performing snapshot.

\subsection{Real-world 3D lane dataset}

Acquiring ground truth labeled data with 3D for the task is an endeavor that requires a complex multiple-sensor setup and possibly also expensive HD maps. To this end we introduce a new such dataset, \textit{3D-lanes}, created using a multi-sensor setup including a forward-looking camera, a Velodine HDL32 lidar scanner and a high-precision IMU all synchronized and accurately aligned. The data was collected in 6 drives each on a different road segment totaling nearly 2 hours of driving. Using the Lidar and IMU we generate aggregated lidar top view images as in~\cite{Urmson_boss}, which are then used together with a semi-manual annotation tool for generating ground truth. In total, $85K$ images were annotated, out of which $1K$, consisting of a separate drive, were used as the test set and the remaining as the train set. The lidar information is additionally used to provide the full 3D curve of each lane. A disadvantage of this approach is that lanes not sufficiently visible to the lidar, due to occlusions or limited resolution at distance, are missing from the ground truth. Therefore, the labeling is somewhat noisy as can be observed in Fig.~\ref{fig:results}(Right column). In addition, the dataset variability in terms of geometry and topology is modest compared to the \emph{synthetic-3D-lanes} dataset. We therefore used the synthetic data which has perfect ground truth to develop the method and conduct ablation studies while the real-world dataset is used for validating transferability of the approach to real data and qualitative analysis. 

\subsubsection{Evaluation results}

\textbf{Evaluation metrics.} We propose an evaluation of 3D lane detection that separates the detection accuracy from the geometric estimation accuracy. \textit{Detection accuracy} is computed via the standard average precision (AP) measure of the precision-recall curve. We first compute a curve to curve distance between a GT and a detected lane as a weighted sum of point-wise Euclidean distances. We measure distances on a set of predefined y-values along the curves, every 80cm in the range 0-80 meters. Weight is decreased for farther away points. We then perform a one-to-one (curve) matching by choosing pairs in decreasing similarity. A matching is considered correct if the weighted distance is below a certain, rather permissive, threshold (1.5 meters). Iterating over lane confidence thresholds we generate the precision-recall curve.

For matched detections we assess the \textit{geometric estimation accuracy} by measuring the distribution of the error (point-wise Euclidean distance) over the same points used to measure the curve to curve distance. We further divide the entire dataset to lane points in the \textbf{near range} (0-30m) and \textbf{far range} (30-80m) due to the differences in the magnitude of errors. We then compute per range the $1\sigma$ error, as the $68$ error percentile and the $2\sigma$ error as the $95$ percentile. Lane centerline and delimiter detection are separately evaluated using this methodology. Irrelevant lane points are ignored in the evaluation as in the training phase.

\textbf{Implementation details.} 3D-LaneNet and all the variants brought in this section were initialized and trained using an identical protocol. The image-view pathway is initialized from VGG16 trained on imagenet~\cite{imagenet_cvpr09}. We train with Adam optimization~\cite{adam} and with initial learning rate $5\cdot 10^{-4}$. We use a variation on the cyclic learning rate regime described in~\cite{cyclic_lr} with a minimal learning rate of $10^{-6}$. The y-range of the top view representation is 80 meters and the x-range is 20 meters. The IPM scale is different in x and y: in the first top-view feature map each pixel corresponds to 16cm laterally (x) and 38.4cm longitudinally (y). The last top-view feature map is $\times 8$ smaller and since there is an anchor per column the distance between anchors is $16\times8=128cm$. We set the $K(=6)$ vertical reference points to be $\mathbf{y}=\left\lbrace5,20,40,60,80,100 \right\rbrace$ and $Y_{ref}=20m$. 

\textbf{Results on \textit{synthetic-3D-lanes} dataset.} Typical network results on the test set are shown in Figure~\ref{fig:results}, with ground truth marked. The first row in Table \ref{tab:ablation} shows the quantitative results of 3D-LaneNet for centerline detection. A valid concern with synthetic datasets is that its variability is too limited and that the learned network memorizes the entire example space instead of learning to generalize. A positive indication that this is not the case is that test AP (0.952) is well below train AP (0.966) as are the geometric evaluation measures. All networks trained in the ablation tests presented here were initialized from VGG16 just as the 3D-LaneNet was and were trained with the same training parameters and number of epochs.

We first examine the role of the \textit{\textbf{dual-pathway architecture}} by comparing it to alternative architectures. The \textbf{image-view} only version connects the image-view pathway directly to the lane detection head which outputs the representation in $\mathcal{C}_{road}$ exactly as 3D-LaneNet does. The anchor positions $X_A$ in this case are determined by the columns in the last feature map: for each column we pick a pixel at a predefined image y-coordinate and project it to top-view to determine the anchor corresponding to the column. The \textbf{top-view} only version first projects the \textit{image} itself to top view and continues the same computation as the top-view pathway. In addition, we tested two versions which include a limited version of the dual-pathway. The \textbf{early IPM} includes a single dual context module (the first amongst the four in the full network). The \textbf{late IPM} similarly contains only the last dual context module out of the four. The results,  summarized in table~\ref{tab:ablation}, show that the full dual-pathway architecture has superior performance compared to all other variants. In particular, the worst result is delivered by the \textbf{image-view} only version, stressing the importance of the top-view processing pathway. Note that the \textbf{late stage IPM}, consisting of a trimmed version of the dual pathway, delivers the second best accuracy, but with a reduced computational cost, making it a good candidate for real-time implementations.

We also tried alternative \textit{\textbf{definitions of the road projection plane}}. One approach takes into consideration the entire scene when fitting the road plane and not only the local road normal. To test it we devised a ground truth generation algorithm which takes the farthest visible road point and connects it to the local road position to determine the pitch. This method, is termed \textbf{horizon} in Table~\ref{tab:ablation} since it resembles horizon estimation methods. Evidently, it performed slightly worse in general, although we observed consistently cases in which the scene topography favors this definition. We also tried assuming a \textbf{fixed position} of the camera, in which the average pitch ($2.5^\circ$) and camera height (165cm) were used to define $T_{c2r}$. Finally, we note that learning to predict the best road projection plane per scene \textit{without explicit supervision}, as proposed in~\cite{neven}, failed to produce satisfying results for our task. 

The last row in Table \ref{tab:ablation} (\textbf{flat ground}) is brought to stress the importance of full 3D lane estimation compared to the current existing approach: image-only detection and \textit{image-to-world} translation using the flat-ground assumption. Image-only detection is obtained by projecting the 3D-LaneNet results to the image plane. For the image-to-world stage we need to choose the plane to project the image result to. We tried two options, both computed using the ground truth: the road plane $P_{road}$ and the plane defined by \textbf{horizon} as described in the previous experiment. As one may expect, the \textbf{horizon} based method, which essentially uses the best planar fit for the entire scene, produced the better results, which are still inferior to those of 3D-LaneNet which performs full 3D estimation.

The \textit{delimiter detection} performance obtained by 3D-LaneNet is $0.971$ AP (positional errors: 12.9cm@$1\sigma$, 33cm@$2\sigma$ near range;  30cm@$1\sigma$, 106cm@$2\sigma$ far range). These metrics show a slightly better performance compared to centerline detection. A possible explanation is that delimiters are clearly marked on the road while centerlines are indirectly inferred. Since output is transformed from road to camera coordinates using an estimated $T_{c2r}$ we also measured the quality of this estimation and its effect on the results. The median values of the absolute errors for pitch ($\theta$) and camera height ($h_{cam}$) are $0.09^\circ$ and $2.4cm$ respectively. To eliminate the contribution of this error we evaluated performance in road coordinates $C_{road}$ by taking the raw network output (before transforming to $C_{camera}$) and got a negligible difference in measured performance.

\textbf{Results on \textit{3D-lanes} dataset.} For operation on real-world data we trained the 3D-LaneNet on the train part of the \textit{3D-lanes} dataset. Result examples from the respective test set are shown in Fig.~\ref{fig:results} (Right column). Note that since the camera is mounted with a downward pitch the 3D lanes are detected as rising upward. Evaluation metrics are presented in Table~\ref{tab:real}. As in the synthetic data, using the \textbf{flat ground} assumption on the real data degrades performance, achieving a 4 times larger error in the far range.

\setlength{\heavyrulewidth}{1.5pt}
\setlength{\abovetopsep}{4pt}
\begin{table}[!htbp]
	\centering
	\caption{Centerline detection results on \emph{synthetic-3D-lanes} dataset}
	\begin{tabular}
		{*6c}
		\toprule
		& \textbf{AP} &   \multicolumn{2}{c}{Error \textbf{near} (cm)} & \multicolumn{2}{c}{Error \textbf{far} (cm)}   \\ 
		&    & $1\sigma$ & $2\sigma$ & $1\sigma$ & $2\sigma$\\
		\textbf{3D-LaneNet} & \textbf{0.952}	& \textbf{13.3}& 	\textbf{34.4}& 	\textbf{33.1}& 	\textbf{122}\\
		\textbf{image-view} & 0.819	& 20.3 & 50	& 74.7	& 241  \\ 
		\textbf{top-view} & 0.929 &	17.5 &	39.6 & 49.5 & 208  \\ 
		\textbf{early IPM } & 0.934 & 13.7	 & 35.5 & 43.5 & 189 \\ 
		\textbf{late IPM } & 0.948	&14.5	&37.2	& 37.4	&139  \\ 
		\midrule 
		
		\textbf{horizon} &0.949	&14.8&	40.4&	36.7&	132	  \\ 	
		\textbf{fixed position }	& 0.948	& 13.6	& 37.3	& 35.4	& 139  \\ 	
		\midrule 
		\textbf{flat ground}&0.566	& 46.9 & 114	& 99	& 289  \\ 	
		\bottomrule
	\end{tabular} 	\label{tab:ablation}
\end{table}

\setlength{\heavyrulewidth}{1.5pt}
\setlength{\abovetopsep}{4pt}
\begin{table}[!htbp]
	\centering
	\caption{Delimiter detection results on \emph{3D-lanes} dataset}
	\begin{tabular}
		{*6c}
		\toprule
		& \textbf{AP} &   \multicolumn{2}{c}{Error \textbf{near} (cm)} & \multicolumn{2}{c}{Error \textbf{far} (cm)}   \\ 
		&    & $1\sigma$ & $2\sigma$ & $1\sigma$ & $2\sigma$\\
		\textbf{3D-LaneNet} & \textbf{0.918}	& \textbf{7.5}& 	\textbf{19.6}& 	\textbf{12.4}& 	\textbf{33}\\
		\midrule 
		\textbf{flat ground} & 0.894 & 19.1 &	37.4 & 64.1	& 137  \\ 	
		\bottomrule
	\end{tabular} 	\label{tab:real}
\end{table}

\subsection{Evaluation of image-only lane detection}

The purpose of this experiment is to compare our approach to the current state of the art, which exists for image-only lane detection. The tuSimple lane dataset~\cite{tu_simple} consists of 3626 training and 2782 test images. Unfortunately, today there is no access to the labels for the test images. We therefore divide the original training set to our own train/validation sets (90\% train and 10\% validation). While we are aware that there may be deviations between our evaluation (obtained on the validation set) and the one on the test set, we can expect a similar performance and reach the same conclusion qualitatively. Since this dataset does not contain 3D information, we train a variation of 3D-LaneNet, which detects the lanes in the image domain. Instead of a 3D representation, the network output was reduced to 2D points on the road projection plane by eliminating the elevation ($\mathbf{z}_t^i$) component. Only the delimiter output type is maintained ($t=d$) since the marked entities in the dataset are lane delimiters. A fixed homography, $H_{tuSimple}$, between image plane and road projection plane, was manually selected, such that straight lanes become parallel in top view. The lanes directly predicted by the network are transformed to lanes in the image view using $H_{tuSimple}$. Since $H_{tuSimple}$ is fixed, the road projection plane prediction branch is not used. Other than the aforementioned, the network is identical to the 3D-LaneNet as configured for the synthetic-3D-lanes dataset. The tuSimple main evaluation metric (\textit{acc})~\cite{tu_simple} is the average ratio of detected ground truth points per image. Using our end-to-end approach on our validation set we reached an accuracy of $0.951$ which is competitive with the one achieved by the tuSimple 2017 competition winning method~\cite{SCNN}, ($0.965$). This result is encouraging and somewhat surprising given that our entire approach was designed towards the 3D estimation task. In particular, our geometric loss (Eq.~\ref{eq:loss}) is computed in top view coordinates, giving in practice a much higher weight to distant lane points while in the tuSimple $acc$ metric all points equally contribute.


\section{Conclusions}
We presented a novel problem, 3D multiple lane detection, along with an end-to-end learning-based solution, 3D-LaneNet. The approach has been developed using a newly introduced synthetic dataset and validated on real data as well. The approach is applicable in principle to all driving scenarios except for complex urban intersections. Finally, we believe that the \textit{dual-pathway} architecture can facilitate additional on-road important 3D estimation tasks such as 3D vehicle detection.
{
	\small
	\bibliographystyle{ieee_fullname}
	\bibliography{egbib}
}

\clearpage
\newpage

\section*{Appendix I - Synthetic data generation details}

\begin{figure*}
	\begin{center}
		\begin{tabular}{ccccccc}
			\includegraphics[width=0.14\linewidth]{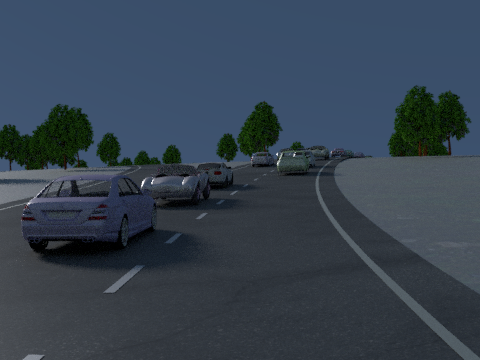} &
			\includegraphics[width=0.14\linewidth]{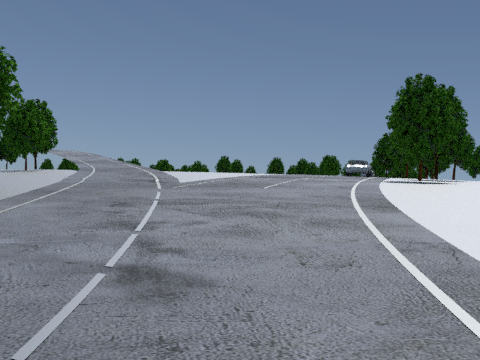} &
			\includegraphics[width=0.14\linewidth]{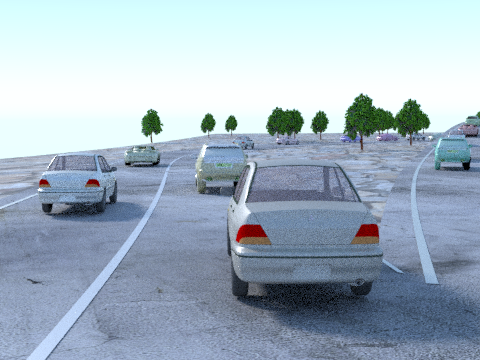} &
			\includegraphics[width=0.14\linewidth]{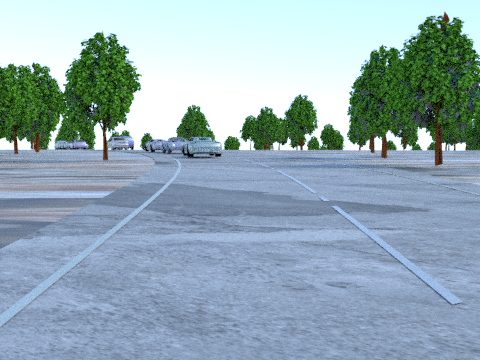} &
			\includegraphics[width=0.14\linewidth]{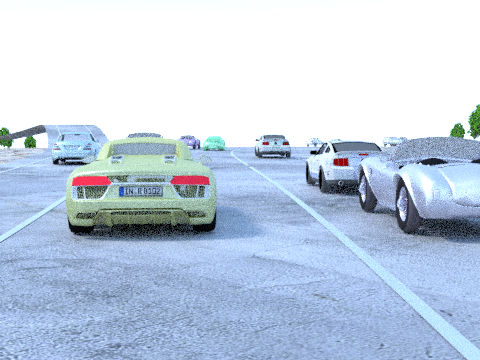} &
			\includegraphics[width=0.14\linewidth]{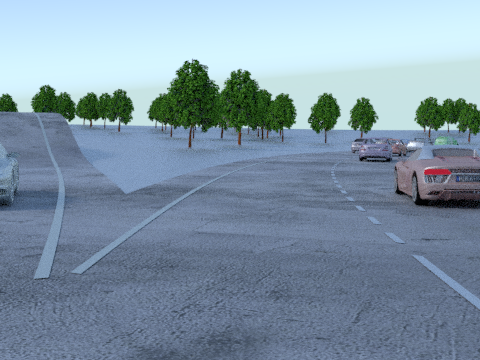} \\
			\includegraphics[width=0.14\linewidth]{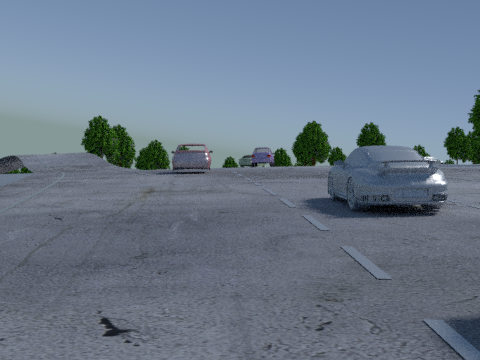} &
			\includegraphics[width=0.14\linewidth]{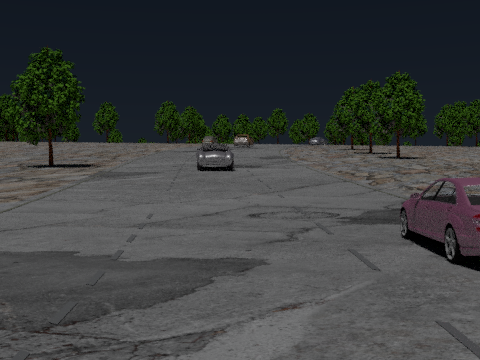} &
			\includegraphics[width=0.14\linewidth]{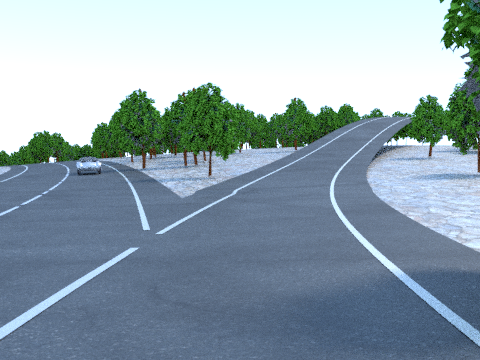} &
			\includegraphics[width=0.14\linewidth]{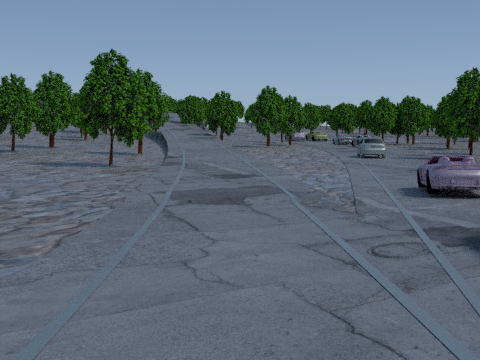} &
			\includegraphics[width=0.14\linewidth]{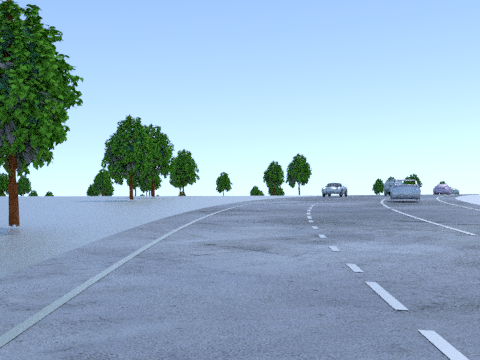} &
			\includegraphics[width=0.14\linewidth]{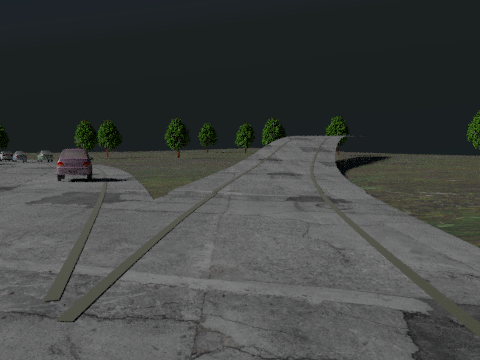} \\
			\includegraphics[width=0.14\linewidth]{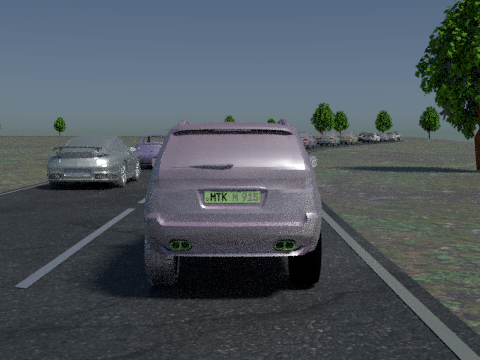} &
			\includegraphics[width=0.14\linewidth]{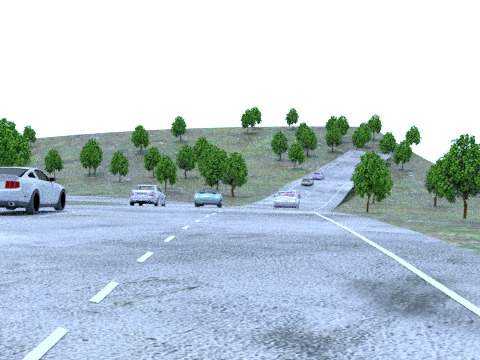} &
			\includegraphics[width=0.14\linewidth]{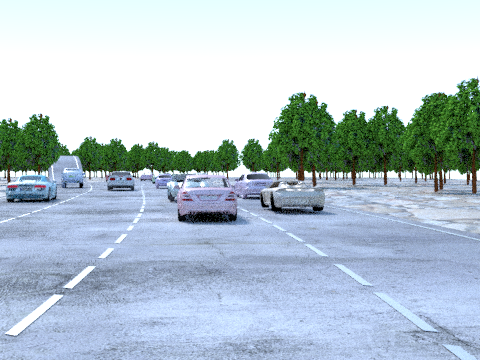} &
			\includegraphics[width=0.14\linewidth]{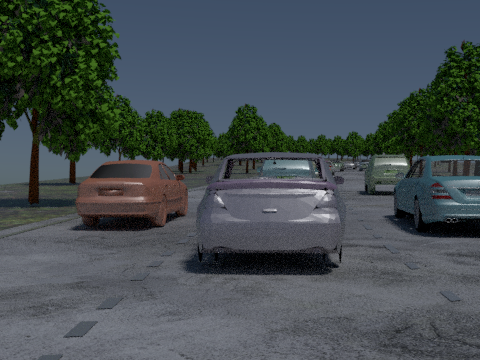} &
			\includegraphics[width=0.14\linewidth]{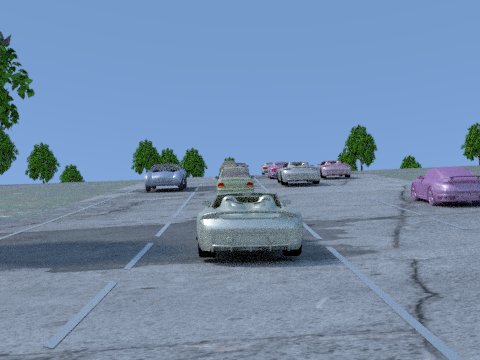} &
			\includegraphics[width=0.14\linewidth]{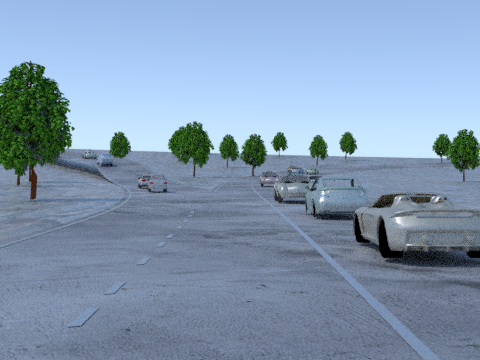} \\
			\includegraphics[width=0.14\linewidth]{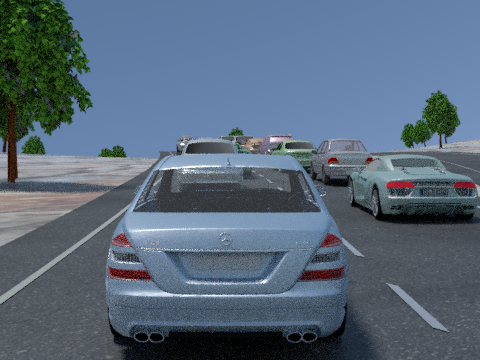} &
			\includegraphics[width=0.14\linewidth]{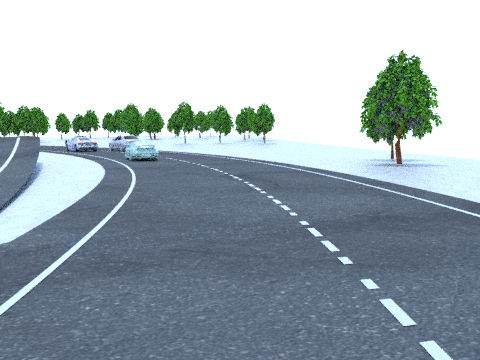} &
			\includegraphics[width=0.14\linewidth]{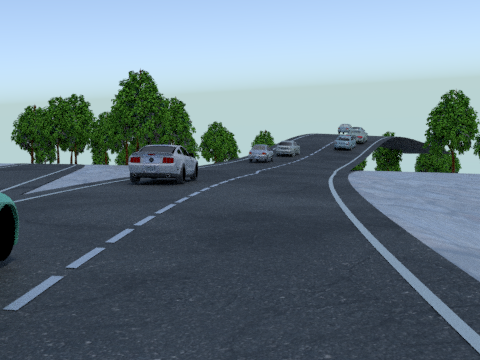} &
			\includegraphics[width=0.14\linewidth]{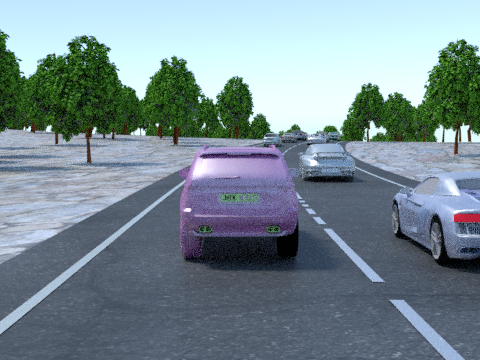} &
			\includegraphics[width=0.14\linewidth]{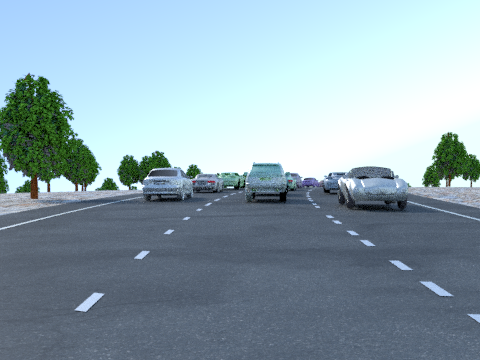} &
			\includegraphics[width=0.14\linewidth]{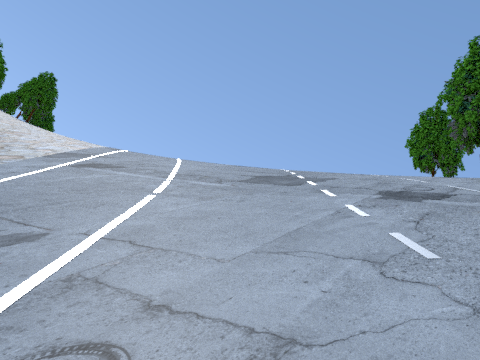} \\
		\end{tabular} 
	\end{center}
	\caption{\textbf{Examples of generated scenes from \textit{synthetic-3D-Lanes}.}}
	\label{fig:examples}
	
\end{figure*}

In this appendix we provide details on the \textit{synthetic-3D-Lanes} dataset generation. As described in Section~\ref{sec:synthetic}, the idea was to generate a large variety of variations in the road and lane topology, topography and curvature, and to introduce natural occuring variations due to occlusions and lighting. Figure~\ref{fig:synthetic} shows an example of a synthetic scene generation of the static elements, and final examples of generated scenes are shown in the left and middle columns of Figure~\ref{fig:results}. Figure~\ref{fig:examples} provides additional examples exemplifying the diversity in all the generating factors, from the geometry of the surface to the lighting and objects placed in the scene. Generating a scene consists of a sequence of random selections as described inSection~\ref{sec:synthetic}.

Tables~\ref{tab:params1}-~\ref{tab:params_last} provide the specific parameters used to generate the dataset. Note that each table corresponds to a stage in the generation process as described in Section~\ref{sec:synthetic} All parameters were uniformly sampled within the specified ranges. The entire world model is built relative to a 3D coordinate system such that the y-axis is roughly aligned with the driving direction, the x-axis is the lateral direction and the z-axis is in the elevation upward direction. The origin (point $(0,0,0)$) is placed in the middle of the scene in top view, and the main road always passes through it. In addition, whenever a secondary road exists (i.e. when splits or merges are modeled), it meets the main road at the origin.

\setlength{\heavyrulewidth}{1.5pt}
\setlength{\abovetopsep}{4pt}
\begin{table*}[h]
	\centering
	\caption{\emph{Synthetic 3D-lanes} dataset parameters: \textbf{Terrain 3D.}}
	\begin{tabular} {*5c}
		\toprule
		\textbf{Parameter} &  \textbf{Min Value} &  \textbf{Max Value} & \textbf{Unit/Type} & \textbf{Description}  \\
		\toprule 
		\#Components   & 1 & 7 & discrete & Terrain is defined by a Gaussian Mixture with \\
		& & & &  this number of components.\\
		\midrule
		Gaussian centers   & -150 & +150 & meters & The Gaussian center in each dimension \\
		& & & & ($x$ and $y$) is chosen within this range.\\
		\midrule
		
		Gaussian magnitude & -50 & 50 & meters & Chosen independently for \\
		& & & & each Gaussian. \\
		\midrule
		Gaussian SD & 25 & 250 & meters & SD=Standard Deviation.\\
		& & & & Chosen independently for each direction (x,y)\\
		\midrule
		Gaussian orientation & 0 & 90 & degrees &\\
		\bottomrule
	\end{tabular} 
	\label{tab:params1}
\end{table*}

\begin{table*}[h]
	\centering
	\caption{\emph{Synthetic 3D-lanes} dataset parameters: \textbf{Road and lane topology.}}
	\begin{tabular} {ccl}
		\toprule
		\textbf{Parameter} &  \textbf{Values} &  \textbf{Description}  \\
		\toprule
		Topology type & 1-4 & 
		\textbf{1. No exit - simple road}\\
		& & \textbf{2. Exit with single lane.} The rightmost lane of the main road splits to \\ 
		& & create and exit and also continues as rightmost lane of the main road. \\
		& & \textbf{3. Exit with single lane II}. The rightmost lane of the main road becomes the exit lane. \\ 
		& & The second rightmost lane of the main road splits to become rightmost and second-right lane.\\
		& & \textbf{4. Exit with two lanes.} The rightmost lane of the main road becomes the right exit lane.\\ 
		& & The second rightmost lane of the main road splits to the left exit lane and \\ 
		& & to the rightmost lane of the main road\\
		\midrule
		Flip longitudinal & Yes/No & Flipping around the longitudinal axis transforms a right split (if exists) into a left one. \\
		
		\midrule
		Flip lateral & Yes/No & Flipping around the lateral axis transforms a split (if exists) into a merge\\
		
		\bottomrule
		
	\end{tabular} 
	\label{tab:real}
\end{table*}

\vspace{5cm}
\begin{table*}[h]
	\centering
	\caption{\emph{Synthetic 3D-lanes} dataset parameters: \textbf{Lane top view geometry in defined (x,y) plane}.}
	\begin{tabular} {ccccl}
		\toprule
		\textbf{Parameter} &  \textbf{Min Value} &  \textbf{Max Value} & \textbf{Unit/Type} & \textbf{Description}  \\
		\toprule		
		\#Lanes on main road & 2 & 4 & & \\
		\midrule
		Lane width & 3.2 & 4 & meters& \\
		\midrule
		Shoulder width & 0.2 & 0.6 & & Factor of lane width \\
		\midrule
		
		Main Road curvature & -10& 10& meters& The geometry of the main road is modeled as \\ 
		& & & & a 4\textsuperscript{th} degree polynomial defined by 5 points: \\
		& & & & $\{(x^{o}_{-50},-50),(x^{o}_{-50}+x^{o}_{-100},-100),(0,0),$\\
		& & & & $(x^{o}_{50},50),(x^{o}_{50}+x^{o}_{100},100)\}$ where each of the \\
		& & & & lateral relative offsets, $x^{o}_{\{-100,-50,50,100\}}$, \\
		& & & & is sampled from the given range. \\
		
		\midrule
		
		Secondary road start angle & 1 & 5 & degrees & Relative to main road at exit point\\
		\midrule
		
		Secondary road curvature & 0  & 10  & meters & Lateral offset 60m after exit.\\
		& & & & Together with the split point $(0,0)$,\\
		& & & &  and the \textit{start angle} define a\\
		& & & &  quadratic polynomial for the secondary road.\\
		\midrule

		Scene boundaries & & & meters & Are set to encompass all roads as defined above\\
		\bottomrule
		
	\end{tabular} 
	\label{tab:real}
\end{table*}
\vspace{5cm}
\begin{table*}[h]
	\centering
	\caption{\emph{Synthetic 3D-lanes} dataset parameters: \textbf{Lane 3D}.}
	\begin{tabular} {*5c}
		\toprule
		\textbf{Parameter} &  \textbf{Min Value} &  \textbf{Max Value} & \textbf{Unit/Type} & \textbf{Description}  \\
		\toprule
		\multicolumn{5}{c}{Note: for the main road 3D is uniquely defined by combining the top-view geometry and the terrain elevation.}	\\	
		\midrule
		Ramp max height	& 2 & 6 &	meters & Ramp height for secondary road\\
		\midrule
		Ramp slope	& 0.5 & 4.5 &	Factor & Together with prev. param defines the ramp length as \\
		& & & & (Ramp max height$\times$Ramp slope) \\
		
		\bottomrule
		
	\end{tabular} 
	\label{tab:real}
\end{table*}

\vspace{5cm}

\begin{table*}[h]
	\centering
	\caption{\emph{Synthetic 3D-lanes} dataset parameters: \textbf{Terrain and Road appearance}.}
	\begin{tabular} {*5c}
		\toprule
		\textbf{Parameter} &  \textbf{Min Value} &  \textbf{Max Value} & \textbf{Unit/Type} & \textbf{Description}  \\
		\toprule		
		Dashed lane cycle len.& 0.5 & 4.5 & meters & Defines dash-to-dash distance\\
		\midrule
		Dash length & 0.3 & 1 & Factor & fraction of cycle length\\
		\midrule
		Lane marker width & 0.1 & 0.15 & meters & \\
		\midrule
		Lane marker grayscale & 0.2 & 1 & Factor & Affects lane visibility. From range $[0,1]$. \\
		Lane marker gloss & 0.5 & 1 & Factor & Blender parameter\\
		\midrule
		Road texture type & 1 & 3 & Type & Selection from possible textures\\
		\midrule
		Road texture scale & 10.0 & 30.0 & Factor & Scales the texture applied to road\\
		\midrule
		Road gloss & 0 & 0.2 & Factor & Blender parameter\\
		\midrule
		Terrain texture type & 1 &  2 & Type & \\
		\midrule
		Terrain texture scale  & 5.0 & 15.0 & \\
		\midrule
		Texture orientation & 0 & 90 & degrees & allows rotation of texture\\
		\bottomrule
		
	\end{tabular} 
	\label{tab:real}
\end{table*}

\begin{table*}[h]
	\centering
	\caption{\emph{Synthetic 3D-lanes} dataset parameters: \textbf{Objects}.}
	\begin{tabular} {*5c}
		\toprule
		\textbf{Parameter} &  \textbf{Min Value} &  \textbf{Max Value} & \textbf{Unit/Type} & \textbf{Description}  \\
		\toprule		
		\# of cars & 1 & 24 & & Positioned randomly in lanes\\
		\midrule
		Car model type & 1 & 6 & Type & Model selected per car\\
		\midrule
		Car scaling & 0.9 & 1.1 & Factor & Scales car model size\\
		\midrule
		Car color & $[0,0,0]$ & $[1,1,1]$ & RGB & \\
		\midrule
		Car gloss & 0.3 & 1 & Factor & Blender parameter\\
		\midrule
		\# of trees & 40 & 800 & & Positioned randomly on terrain.\\
		\bottomrule
		
	\end{tabular} 
	\label{tab:real}
\end{table*}

\begin{table*}[h]
	\centering
	\caption{\emph{Synthetic 3D-lanes} dataset parameters: \textbf{Scene rendering}.}
	\begin{tabular} {*5c}
		\toprule
		\textbf{Parameter} &  \textbf{Min Value} &  \textbf{Max Value} & \textbf{Unit/Type} & \textbf{Description}  \\
		\toprule		
		Host car lane & 1 & \#lanes & & On main road\\
		\midrule
		Host car position &  &  & & Position within lane is chosen within limits\\
		& & & & such that viewing direction is towards origin\\
		\midrule
		Host car offset & 0 & 0.4 & meters & Offset from lane center\\
		\midrule
		Camera height& 1.4 & 1.9 & meters & \\
		\midrule
		Camera pitch& 0 & 5 & degrees & Downwards\\
		\midrule
		Sun position in sky & 0 & 45 &  degrees & From zenith, to any xy direction\\
		\midrule
		Scene exposure & 1  & 3 & Factor  & Blender render exposure.\\
		\bottomrule
	\end{tabular} 
	\label{tab:params_last}
\end{table*}

\end{document}